\documentclass[lettersize,journal]{IEEEtran}
\usepackage{cite}
\usepackage{amsmath,amssymb,amsfonts,mathtools}

\usepackage[ruled, linesnumbered]{algorithm2e}
\usepackage{algorithmic}
\usepackage{xcolor}

\usepackage{graphicx}
\usepackage[caption=false,font=small,labelfont=sf,textfont=sf]{subfig} 
\usepackage{array}
\usepackage{booktabs}
\usepackage{multirow}
\usepackage{adjustbox}
\usepackage{makecell}

\usepackage{textcomp}
\usepackage{stfloats}
\usepackage{url}
\usepackage{verbatim}
\usepackage{bm}

\begin{document}

\title{Integrating Clinical Knowledge Graphs and Gradient-Based Neural Systems for Enhanced Melanoma Diagnosis via the 7-Point Checklist}

\author{Yuheng Wang, Tianze Yu, Jiayue Cai, Sunil Kalia, Harvey Lui, Z. Jane Wang, and Tim K. Lee%

\thanks{This work was supported in part by the National Sciences and Engineering Research Council of Canada (2017-04932, RGPIN-2022-03049, 2024-06047), the National Natural Science Foundation of China (62201357), the Shenzhen Science and Technology Program (RCBS20221008093347105), the Medicine Plus Program of Shenzhen University (2024YG006), and the Shenzhen University (SZU) Top Ranking Project Grant (86000000210). The authors also acknowledge contributions from Vancouver Coastal Health. Additionally, this study was supported in part through the computational resources and services provided by Advanced Research Computing at the University of British Columbia. (Corresponding author: Jiayue Cai)}

\thanks{Yuheng Wang, Sunil Kalia, Harvey Lui, and Tim K. Lee are with the Department of Dermatology and Skin Science, the Photomedicine Institute, the Centre for Clinical Epidemiology and Evaluation, the School of Biomedical Engineering, and the Department of Electrical and Computer Engineering, The University of British Columbia, and Vancouver Coastal Health Research Institute, Vancouver, BC, Canada. Yuheng Wang, Sunil Kalia, and Tim K. Lee are also with the Department of Population Health Sciences, BC Cancer, Vancouver, BC, Canada. Sunil Kalia is also with the BC Children’s Hospital Research Institute, Vancouver, BC, Canada (email: yuhengw@ece.ubc.ca; sunil.kalia@ubc.ca; harvey.lui@ubc.ca; tim.lee@ubc.ca).

Jiayue Cai is with the Guangdong Key Laboratory of Biomedical Measurements and Ultrasound Imaging, School of Biomedical Engineering, Shenzhen University Medical School, Shenzhen University, Shenzhen, Guangdong, China (email: jiayuec@szu.edu.cn).

Z. Jane Wang and Tianze Yu are with the Department of Electrical and Computer Engineering, The University of British Columbia, Vancouver, BC, Canada (email: tianzey@ece.ubc.ca; zjanew@ece.ubc.ca).}}

\markboth{August, 2025}%
{Shell \MakeLowercase{\textit{et al.}}: A Sample Article Using IEEEtran.cls for IEEE Journals}


\maketitle

\begin{abstract}
The 7-point checklist (7PCL) is a widely used diagnostic tool in dermoscopy for identifying malignant melanoma by assigning point values to seven specific attributes. However, the traditional 7PCL is limited to distinguishing between malignant melanoma and melanocytic Nevi, and falls short in scenarios where multiple skin diseases with appearances similar to melanoma coexist. To address this limitation, we propose a novel diagnostic framework that integrates a clinical knowledge-based topological graph (CKTG) with a gradient diagnostic strategy featuring a data-driven weighting system (GD-DDW). The CKTG captures both the internal and external relationships among the 7PCL attributes, while the GD-DDW emulates dermatologists' diagnostic processes, prioritizing visual observation before making predictions. Additionally, we introduce a multimodal feature extraction approach leveraging a dual-attention mechanism to enhance feature extraction through cross-modal interaction and unimodal collaboration. This method incorporates meta-information to uncover interactions between clinical data and image features, ensuring more accurate and robust predictions. Our approach, evaluated on the EDRA dataset, achieved an average AUC of 88.6\%, demonstrating superior performance in melanoma detection and feature prediction. This integrated system provides data-driven benchmarks for clinicians, significantly enhancing the precision of melanoma diagnosis.
\end{abstract}


\section{Introduction}
\IEEEPARstart{S}{kin} cancer is a widespread malignancy worldwide, with melanoma being the most dangerous type. Although the incidence of skin cancer varies by region and individual risk factors, timely diagnosis and prompt intervention can significantly improve the prognosis of the disease \cite{me2024global}.

In the medical field, healthcare professionals utilize diverse imaging modalities to expedite the diagnosis and treatment of skin lesions \cite{heibel2020review}. Two notable techniques are dermoscopic imaging and clinical photography. Dermoscopy, which is an optical magnification technique assisted by liquid immersion or cross-polarized illumination, produces intricate images revealing subepidermal structures and lesions to support the diagnosis \cite{celebi2019dermoscopy}. On the other hand, clinical images efficiently capture essential morphological features including color, texture, shape, and borders, despite their inability to explore intricate subcutaneous lesion details. The convenience and accessibility of clinical images, especially through widely available mobile electronic devices like smartphones, are indisputable benefits \cite{wen2022characteristics}. The fusion of these two imaging techniques provides a comprehensive information essential in practical medical scenarios, specifically for melanoma diagnosis.

\begin{figure}[!t]
    \centering
    \subfloat[]{
        \includegraphics[width=\columnwidth]{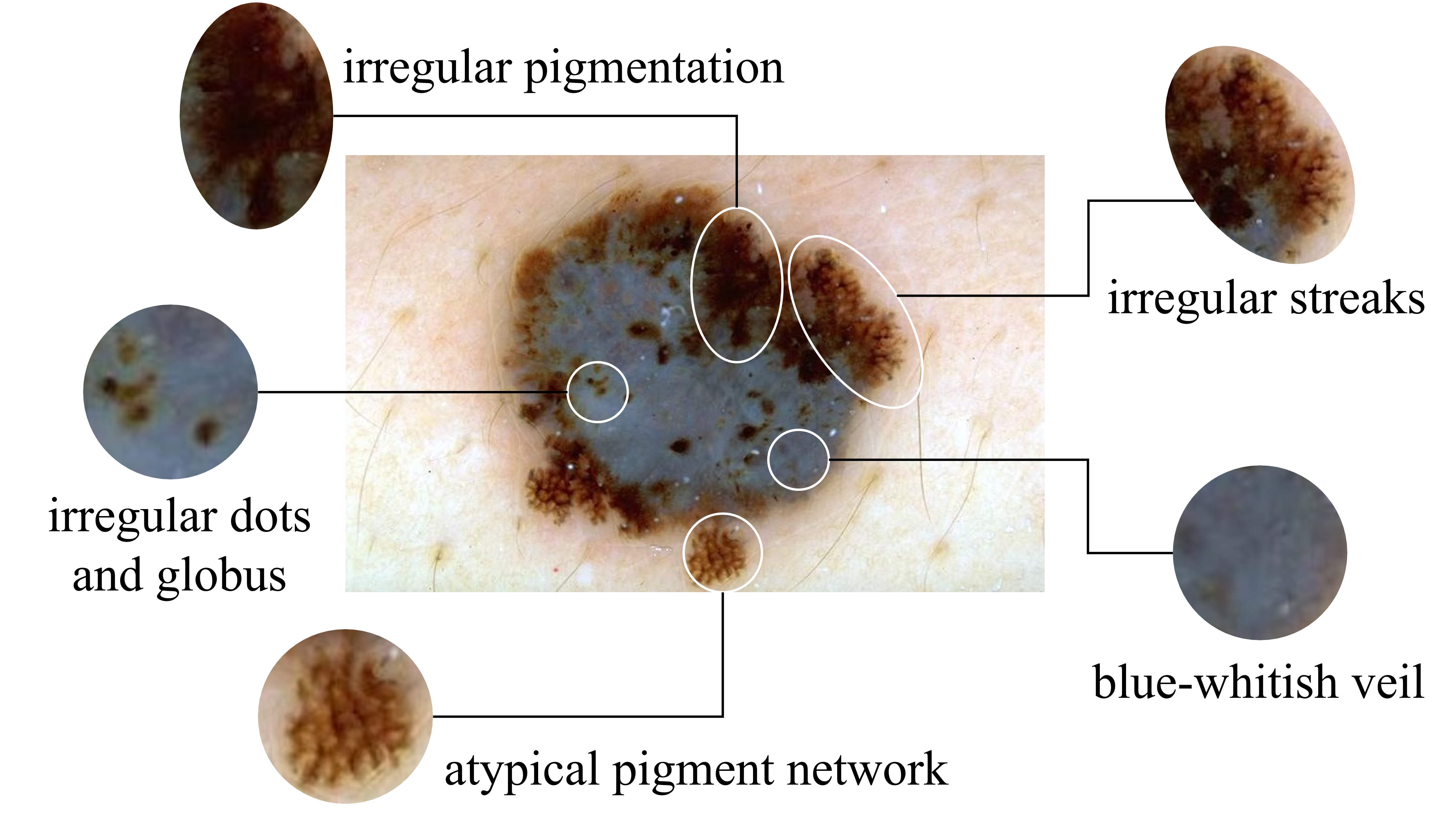}%
        \label{fig:subfig1}
    }
    
    \subfloat[]{
        \includegraphics[width=\columnwidth]{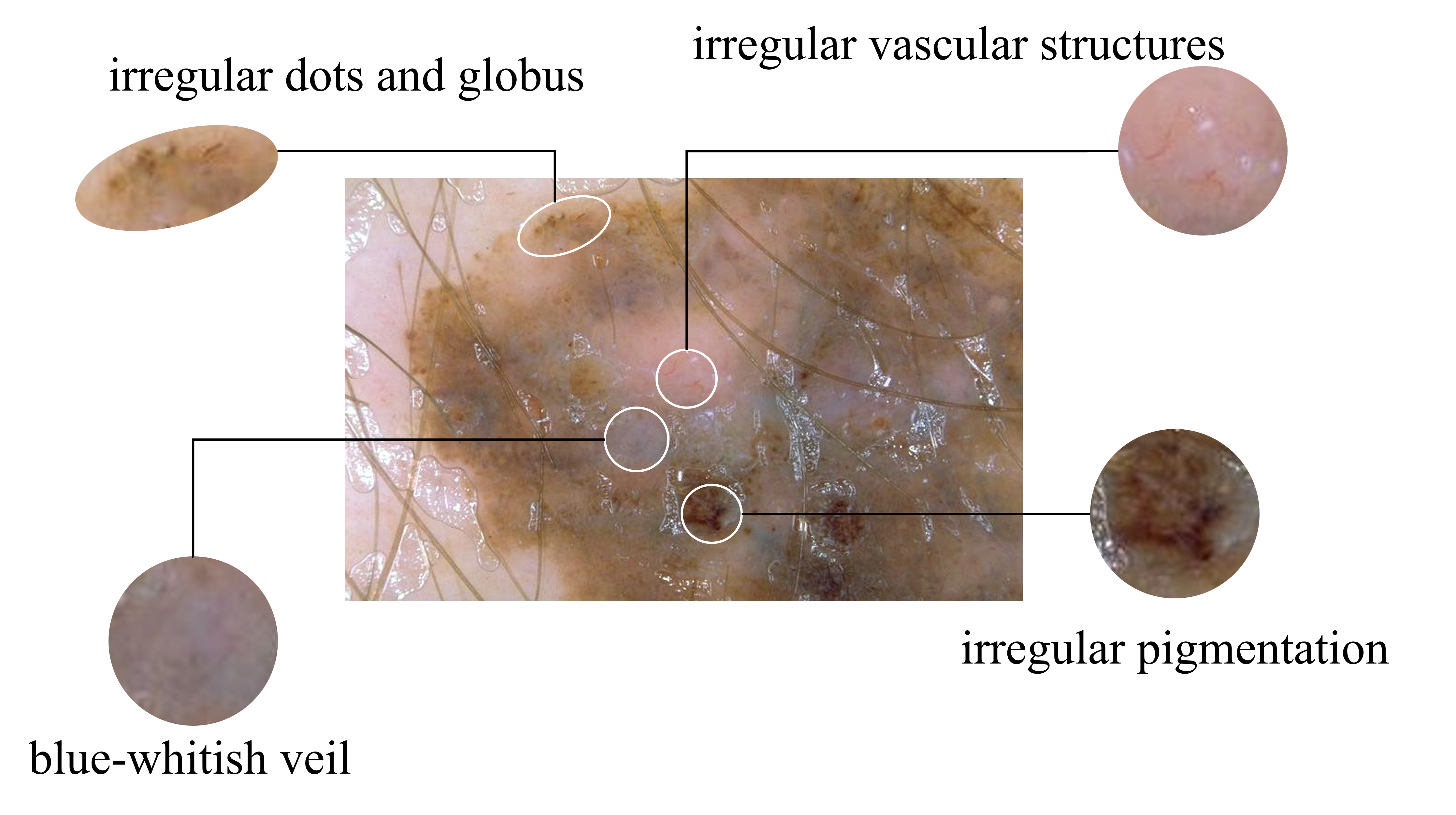}%
        \label{fig:subfig2}
    }
    
    \subfloat[]{
        \includegraphics[width=\columnwidth]{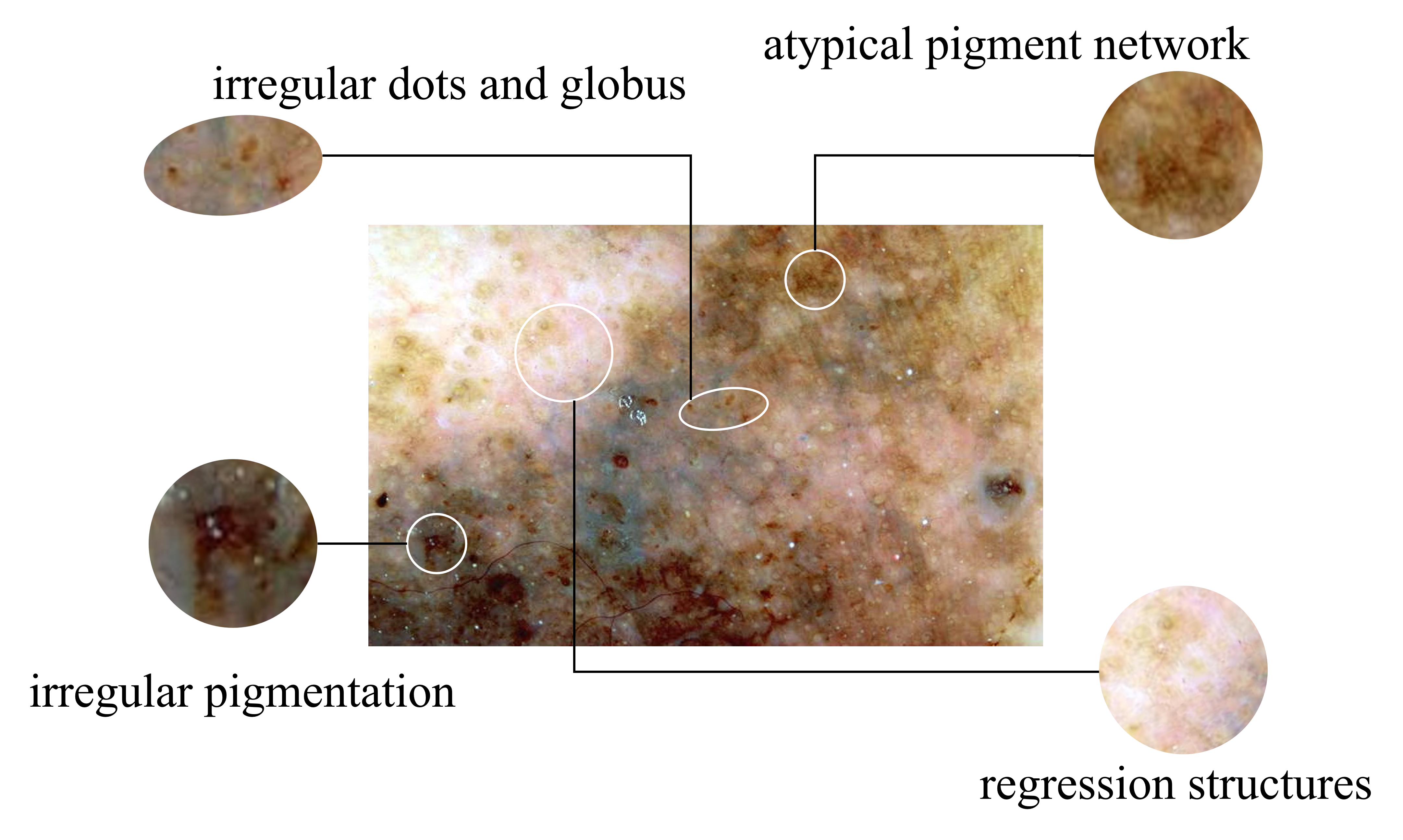}%
        \label{fig:subfig3}
    }
    
    \caption{Example of biopsy-confirmed melanoma cases from the EDRA dataset, annotated with attributes from the 7-point checklist: 
    (a) IR-PIG (1), IR-STR (1), IR-DaG (1), BWV (2), and ATP-PN (2), totaling 7 points; 
    (b) IR-DaG (1), IR-VS (2), BWV (2), and IR-PIG (1), totaling 6 points; 
    (c) IR-DaG (1), ATP-PN (2), IR-PIG (1), and RS (1), totaling 5 points. All cases exceed the threshold of 3 points.}
    \label{example_data}
\end{figure}

Conventionally, clinical practice relies on pattern analysis, which recognizes dermatologic attributes to aid in skin lesion diagnosis. Several established algorithms, such as the ABCD rule \cite{abbasi2004early} and the 7-point checklist (7PCL) \cite{kawahara2018seven}, aid dermatologists in recognizing cancer-related characteristics, contributing to a simpler diagnostic process. For example, this study focuses on the 7PCL algorithm, which evaluates seven dermoscopic characteristics associated with malignant melanoma. These include three major attributes: atypical pigment network (ATP-PN), blue-whitish veil (BWV), and irregular vascular structures (IR-VS), as well as four minor attributes: irregular pigmentation (IR-PIG), irregular streaks (IR-STR), irregular dots and globus (IR-DaG), and regression structures (RS). Each major attribute receives a weighted score of two points, while minor attributes earn one point each. Cumulative scores provide an initial evaluation, requiring a predetermined threshold—commonly set at one or three points. Scores exceeding this threshold are classified as suspected melanoma, as illustrated in Fig.~\ref{example_data}. Although the 7PCL algorithm is widely accepted and utilized, the attributes and their assigned weights are primarily designed to differentiate between malignant melanoma (Mel) and melanocytic Nevi (MN) \cite{argenziano1998epiluminescence}, limiting its range of applications. By analyzing these attributes as key features using more appropriate quantitative criteria, there is potential to broaden its applicability and enhance predictive performance, ultimately benefiting healthcare practitioners. 

\begin{figure}[t]
    \centering
    \includegraphics[width=0.9\columnwidth]{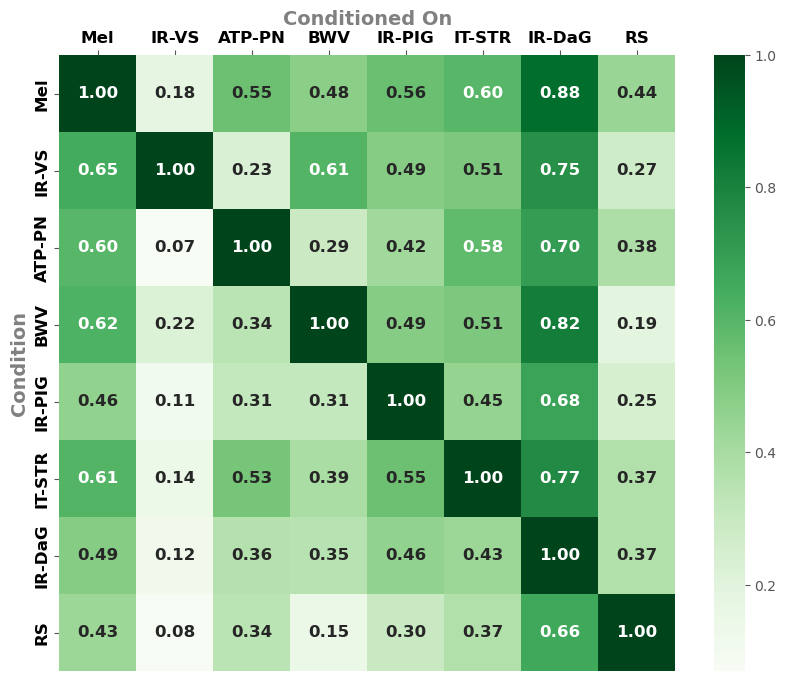}
    \caption{Conditional probability matrix for melanoma and 7PCL attributes. Each cell in the matrix represents the probability of one attribute given the presence of another. The rows correspond to the conditions, and the columns represent the attributes conditioned upon.}
    \label{fig:conditional}
\end{figure}

In this study, we propose a data-driven graph-based framework for melanoma diagnosis that incorporates clinical insights from the 7-point checklist (7PCL). Our work is based on the observation that melanoma attributes exhibit directed relationships, influencing each other to varying degrees, as shown in Fig.~\ref{fig:conditional}. Additionally, the relative importance of these attributes varies depending on the diagnostic context. In summary, the main contributions of this work are:\begin{itemize}
    \item  We develop a directed weighted GCN encoding clinically relevant relationships between melanoma attributes, transforming expert knowledge into a structured, data-driven diagnostic system.
    \item  A dual-attention mechanism enables an effective fusion of image features and meta-information, improving feature extraction and predictive performance.
    \item  Our method assigns trainable weights to 7PCL attributes, ensuring adaptability across different dermatological conditions while preserving the clinical decision logic.
    \item  Our method is evaluated in three open access datasets, EDRA, ISIC2017, ISIC2018, demonstrating improved melanoma prediction and attribute recognition.
\end{itemize}


\section{Related Work}

Computer-aided diagnosis (CAD) has rapidly emerged as an effective tool to assist dermatologists in the detection of melanoma, as it can extract characteristic information related to the lesion. The conventional CAD procedure for melanoma diagnosis based on machine learning involves multiple steps. These include image preprocessing (e.g., artifact removal and color correction), lesion segmentation, extraction of hand-crafted feature sets, and classification using dedicated classifiers based on the application context \cite{barata2017development,wadhawan2011implementation}. Initially, Celebi {\textit{et al.}} \cite{celebi2007methodological} introduced a method that automatically detects borders, extracts shape, color, and texture features from relevant regions, selects optimized features, and addresses class imbalance specifically in dermoscopy images of pigmented skin lesions. Based on this work, researchers have proposed ways to take advantage of the variety of hand-crafted features of melanoma. Schaefer {\textit{et al.}} \cite{schaefer2014ensemble} proposed an ensemble method that combines multiple classifiers and various hand-crafted features and demonstrated statistically improved recognition performance. Fabbrocini {\textit{et al.}} \cite{fabbrocini2014automatic} proposed a framework that integrates hand-crafted features and the 7-point checklist. The authors employed both machine learning classifiers and statistical analysis to enhance the performance of diagnosis. Wadhawan {\textit{et al.}} \cite{wadhawan2011implementation} devised a comparable notion and modified it for intelligent, portable devices aimed at routine clinical skin monitoring or as an aid for dermatologists and primary care physicians in identifying melanoma. These efforts have facilitated collaboration between machine learning methods and clinical diagnosis of melanoma. However, their performance has been limited by an over-reliance on preprocessing and complicated feature extraction procedures.

Over the past decade, deep learning has become a widely used and powerful tool for feature learning and pattern recognition in CAD, enabling automated detection of melanoma and other skin diseases in various scenarios using different methods \cite{esteva2017dermatologist,wang2021deep,abhishek2021predicting,barata2023reinforcement}. Esteva {\textit{et al.}} \cite{esteva2017dermatologist} demonstrated the classification of skin lesions using a single convolutional neural network trained end-to-end directly from both clinical and dermoscopy images, using only pixels and disease labels as input, and achieving performance comparable to dermatologists. Then, Abhishek {\textit{et al.}} \cite{abhishek2021predicting} presented the first work to directly explore image-based predicting clinical management decisions of melanoma without explicitly predicting diagnosis. Meanwhile, the International Skin Imaging Collaboration (ISIC) challenges and the corresponding constantly updated and expanded public datasets have greatly contributed to the rapid development of the field, and more excellent methods and training strategies have been proposed \cite{codella2018skin}. However, the above-mentioned research works focus more on the pixel information brought by the image itself, especially in the imaging modality of dermoscopy. Incorporating expert domain knowledge and multimodal data are important directions that deserve further attention.

Using information from clinical diagnostics to support multi-modal deep learning models has proven effective in several related research. Moura {\textit{et al.}} \cite{moura2019abcd} combined the ABCD rules and a pre-trained CNNs through a multi-layer perceptron classifier for melanoma detection and achieved improved performance. Kawahara {\textit{et al.}} \cite{kawahara2018seven} proposed a multitask Convolutional Neural Network (CNN) trained on both dermoscopy and clinical images, along with the corresponding 7PCL attributes for melanoma detection, which has been regarded as the benchmark of the related research studies. Following this work, Bi {\textit{et al.}} \cite{bi2020multi} proposed a hyper-connected CNN (HcCNN) structure for the multi-modal analysis of melanoma and the corresponding clinical features. Tang  {\textit{et al.}} \cite{tang2022fusionm4net} further developed this idea and fused 7PCL attributes information into a deep learning framework in a multi-stage manner, improving the average diagnostic performance. However, none of these studies have fully considered the intrinsic connections between melanoma attributes recorded as meta-information when analyzing the clinical data. The meta-information is processed consistently as feature vectors in combination with image features, resulting in some improvement in the outcomes. Nevertheless, the potential value of this clinical information warrants further investigation. To address these issues, Graph Convolutional Neural Networks (GCNs) are introduced to assist with data mining for works involving supplementary clinical information \cite{rossi2018deep,fu2022graph,gao2023medical}. Wu {\textit{et al.}} \cite{wu2020learning} introduced GCNs based on general clinical meta-information to aid in the multi-label classification of skin diseases. Wang {\textit{et al.}} \cite{wang2021incorporating} designed a framework utilizing constrained classifier chains, and first examined the mutual relationships between attributes of 7PCL. Fu {\textit{et al.}} \cite{fu2022graph} applied GCNs to a 7PCL-based multimodal and multilabeled classification task, and uncovered the interconnections between attributes by analyzing simultaneous occurrences of different attributes and undirected GCNs. These studies demonstrate the potential of combining graph learning with clinical diagnosis to aid dermatologists in the pre-diagnosis of melanoma. However, a gap persists in understanding the directional relationships among diverse attributes and their practical implications for melanoma diagnosis. 

In recent studies, deep learning algorithms incorporating attention mechanisms have gained prominence and achieved significant success in medical image and signal analysis \cite{he2023transformers, azad2023advances,lv2023meta,lv2023ssagcn}. In the field of dermatology, He {\textit{et al.}} \cite{he2023co} introduced a cross-modality structure that applies attention weight maps learned separately from bimodal images to each other, enhancing prediction performance. Xu {\textit{et al.}} \cite{xu2024remixformer++} developed an architecture that integrates global and local attention mechanisms, mimicking the diagnostic process of dermatologists. Additionally, Xiao {\textit{et al.}} \cite{xiao2024attention} proposed an attention-guided feature reconstruction method utilizing both clinical and ultrasound images, leveraging intra- and inter-modality information to extract more discriminative features. While these studies have significantly enhanced the performance of deep learning models for dermatological image analysis and improved the interpretability of their results, they primarily focus on interactions among learned image features alone. The integration of meta-information, such as the 7PCL attributes, and its interaction with image features remains underexplored. Furthermore, the potential for AI technology to provide feedback on the 7PCL and thereby benefit clinical decision-making has yet to be addressed.

\begin{figure*}[t]
    \centering
    \includegraphics[width=\textwidth]{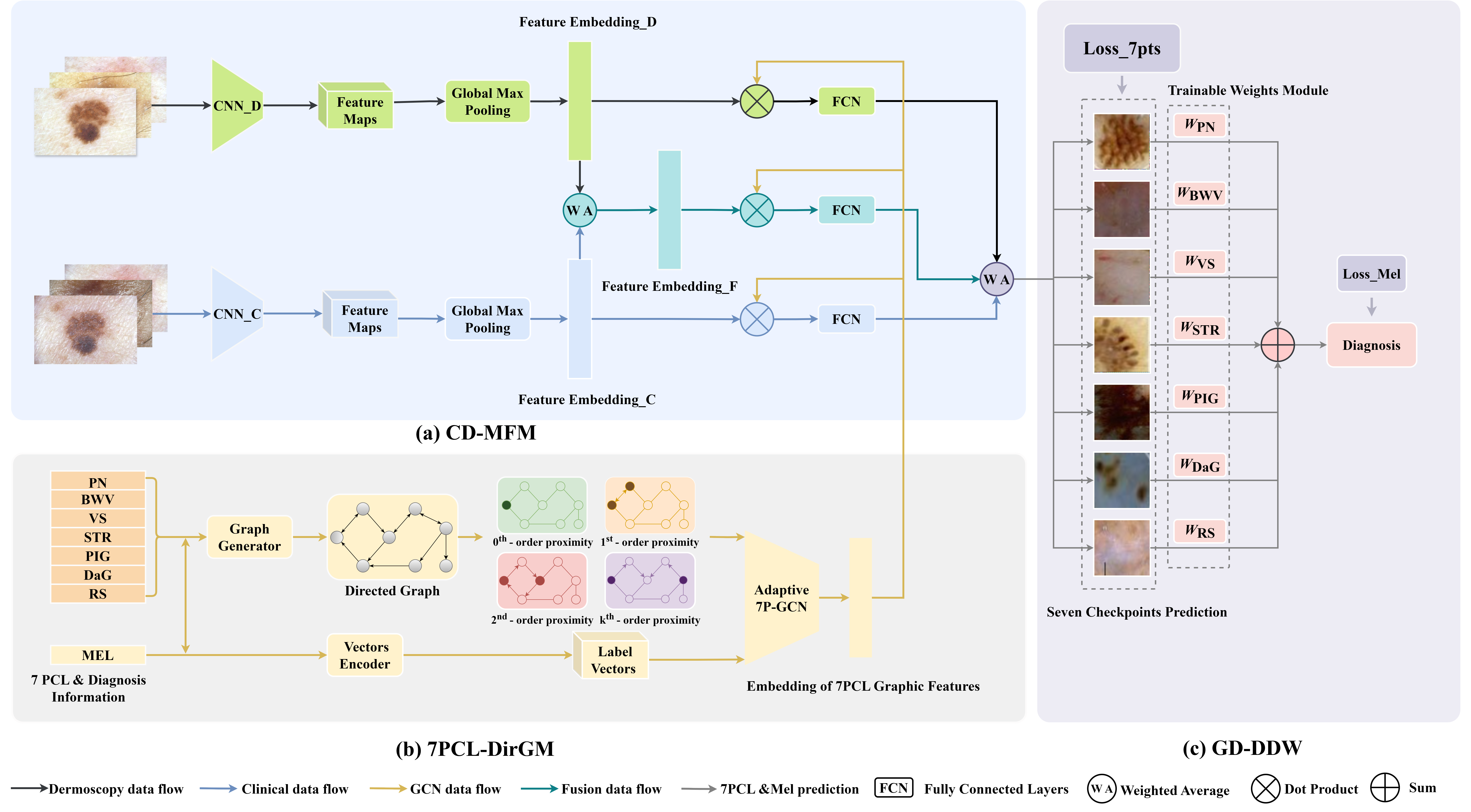}
    \caption{Illustration of our proposed melanoma diagnosis framework, featuring a topological graph convolutional neural network inspired by clinical knowledge and a gradient diagnosis mechanism. The framework includes: (a) the multimodal fusion system combining data from dermoscopy and clinical images, and the meta information; (b) generation of graph features from the 7PCL attributes; and (c) the gradient diagnostic strategy with a trainable weights module.}
    \label{fig:workflow}
\end{figure*}

\section{Methodology}

As illustrated in Fig.~\ref{fig:workflow}, the proposed method comprises three modules. The first module, Clinical-Dermoscopic Multimodal Fusion (CD-MFM), combines information from clinical and dermoscopic modalities. The second module, 7-Point Checklist Directed Graph Mining (7PCL-DirGM), extracts representative features using a mining strategy that leverages directed and multi-order graph information from multi-labeled data. These two modules form our Clinical Knowledge-Based Topological Graph Convolutional Network (CKTG). Additionally, the GD-DDW employs data-driven weighted gradient diagnosis to enhance diagnostic performance by emulating a dermatologist's diagnostic behavior. The following subsections provide a detailed overview of the methodology.

\begin{figure}[t]
    \centering
    \includegraphics[width=0.9\columnwidth]{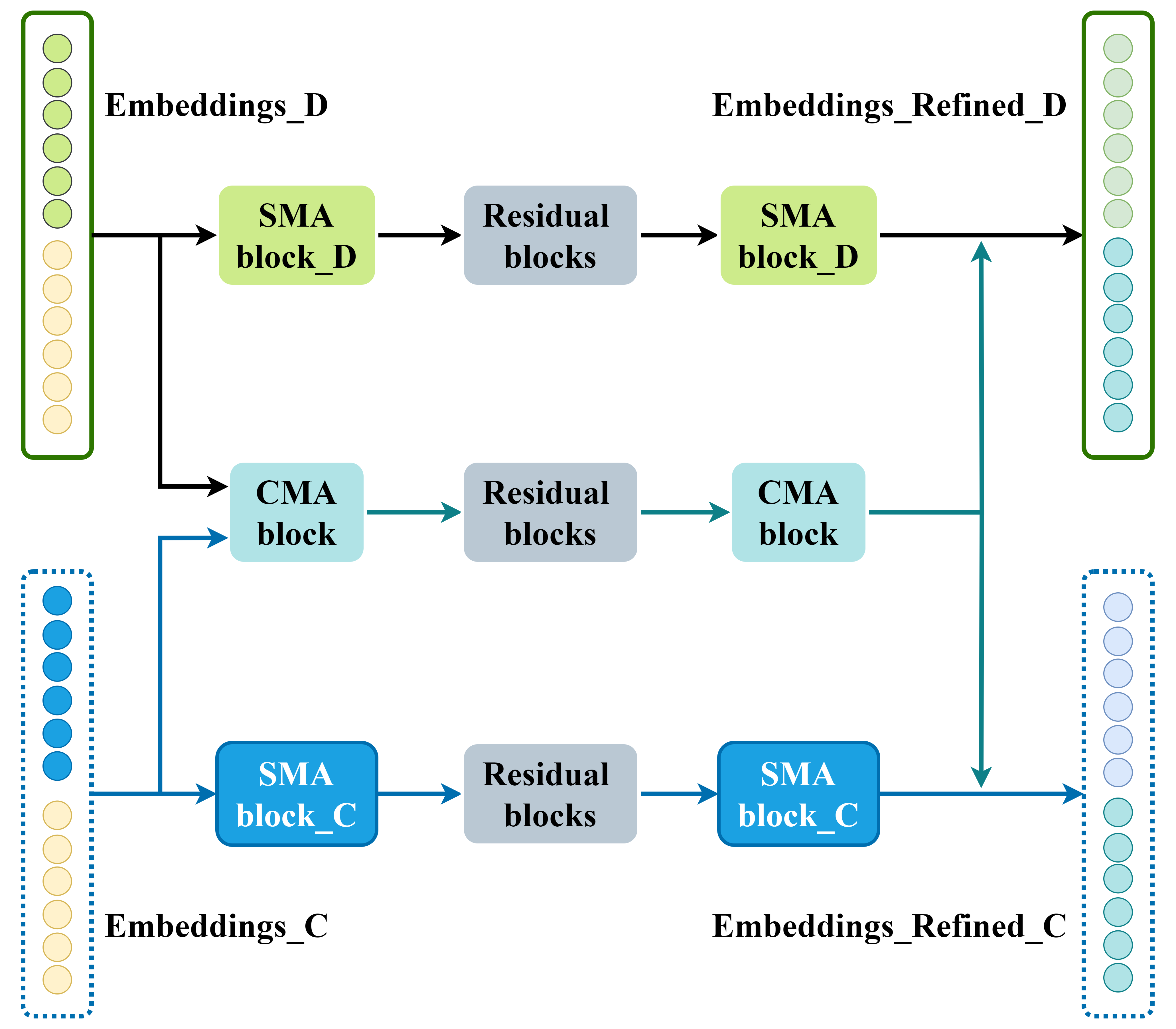}
    \caption{Illustration of the dual-attention blocks}
    \label{fig:dual-attention blocks}
\end{figure}

\subsection{Clinical-Dermoscopic Multimodal Fusion (CD-MFM)}

In this study, we treated each skin lesion as an individual ``case'' within the dataset. Each case, denoted as ${x}^i$, encompasses data from multiple modalities, including dermoscopy images ${x}^i_\textrm{d}$, clinical images $x^i_\textrm{c}$, and the encoded information related to the $j_{\textrm{th}}$ attribute within the 7PCL, $x^i_{\textrm{7p}_j}$, where $i$ ranges from 1 to $n$ (the total number of cases) and $j$ ranges from 1 to 7. Additionally, each case is associated with a diagnostic tag, denoted as $y^i_\textrm{m}$, which represents the melanoma diagnosis, and labels $y^i_{\textrm{7p}_j}$ for their 7PCL attributes.

First, residual convolutional blocks are applied to independently extract visual features from dermoscopic and clinical images, denoted as $X_{\textrm{d}}$ and $X_{\textrm{c}}$, respectively. Simultaneously, a BERT-based semantic embedding layer is used to encode the label information into semantic features $X_{\textrm{m}}$, which represent the diagnosis and 7PCL attributes. These semantic features are aligned with the visual features from the two image modalities. Next, two separate concatenation layers are applied to combine the visual features and the semantic features, resulting in $X_{\textrm{dm}}$ for the dermoscopic modality and $X_{\textrm{cm}}$ for the clinical modality. These combined features are then passed through dual-attention feature extraction blocks for further processing.

As shown in Fig.~\ref{fig:dual-attention blocks}, the dual-attention feature extraction blocks consist of both Single-Modality Attention (SMA)~\cite{fu2019dual} and Cross-Modality Attention (CMA) blocks~\cite{he2023co}. The features $X_{\textrm{dm}}$ and $X_{\textrm{cm}}$ are first refined by their respective SMA blocks, which capture intra-modality interactions within the dermoscopic and clinical image embeddings, \text{embedding\_D} and \text{embedding\_C}, respectively, along with their meta-information in \text{embedding\_m}. These features then pass through residual blocks before being processed by the CMA block, which facilitates cross-modality interaction between the dermoscopic and clinical features. This enables the integration of complementary information from both modalities, enhancing the joint feature representation.

\begin{figure}[t]
    \centering
    \subfloat[SMA Block]{\includegraphics[width=0.95\columnwidth]{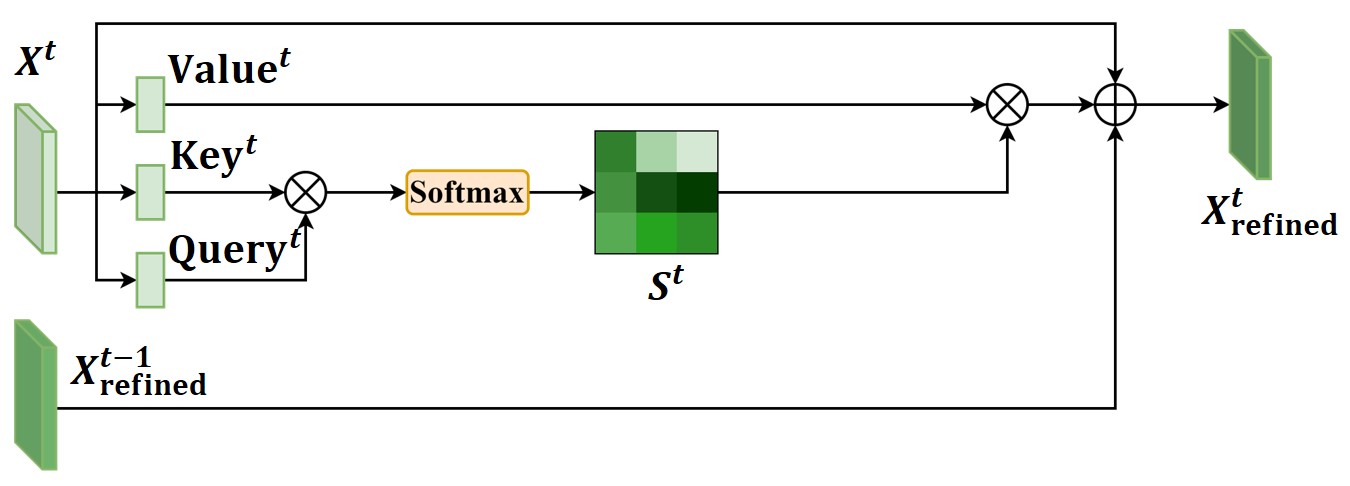}%
    \label{fig:SMA}}
    
    \subfloat[CMA Block]{\includegraphics[width=0.95\columnwidth]{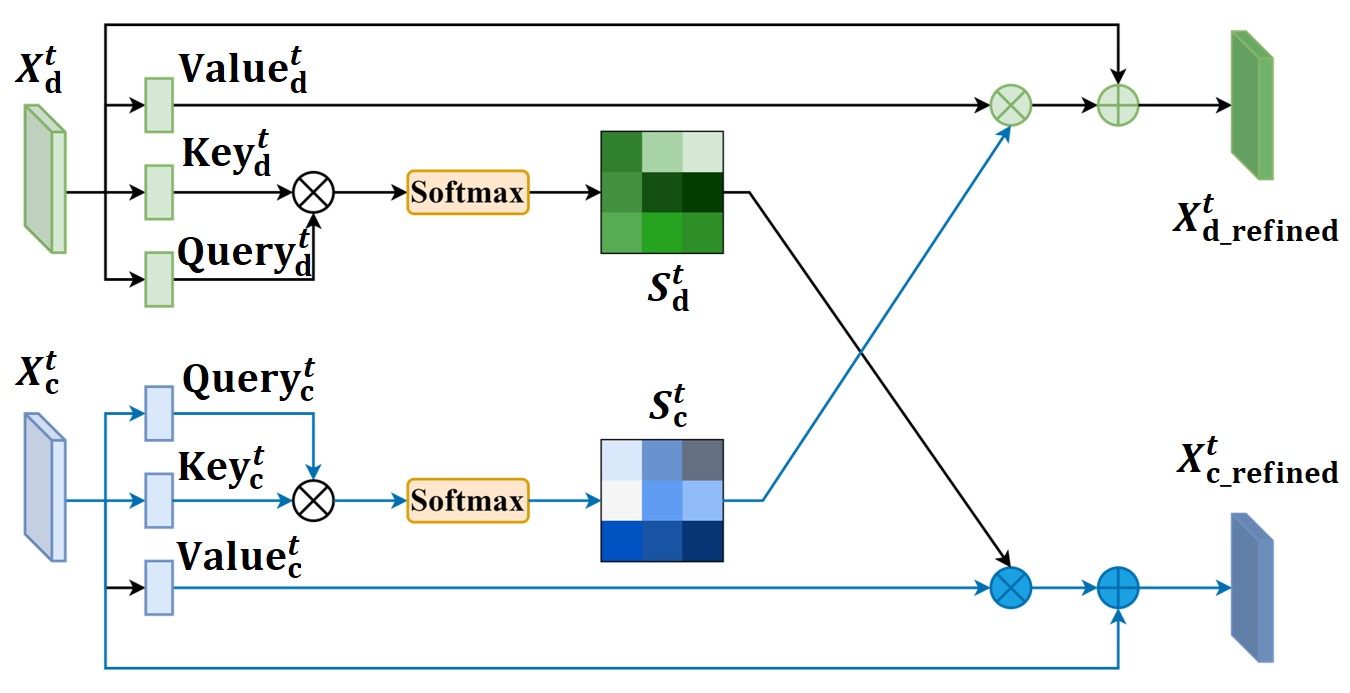}%
    \label{fig:CMA}}
    
\caption{The illustration of Single-Modality Attention (SMA) block and Cross-Modality Attention (CMA) block. The operator $\oplus$ denotes element-wise addition, while $\otimes$ represents element-wise multiplication.}

    \label{fig:CMA_SMA}
\end{figure}

In detail, as shown in Fig.~\ref{fig:CMA_SMA}, the input features $X_{\textrm{dm}}$ and $X_{\textrm{cm}}$ are transformed into query ($Q$), key ($K$), and value ($V$) representations for each modality. The attention scores for each modality are then computed as:

\begin{equation}
\begin{aligned}
S_{\textrm{dm}} &= \text{softmax}\left(\frac{Q_{\textrm{dm}} \cdot K_{\textrm{dm}}^T}{\sqrt{d_k}}\right) \\
S_{\textrm{cm}} &= \text{softmax}\left(\frac{Q_{\textrm{cm}} \cdot K_{\textrm{cm}}^T}{\sqrt{d_k}}\right)
\end{aligned}
\label{eq:SMA_att}
\end{equation}

The refined features for each modality are then obtained by applying the attention scores to the corresponding value representations in the SMA block:

\begin{equation}
\begin{aligned}
X_{\textrm{SMA\_dm}} &= S_{\textrm{dm}} \cdot V_{\textrm{dm}} \\
X_{\textrm{SMA\_cm}} &= S_{\textrm{cm}} \cdot V_{\textrm{cm}}
\end{aligned}
\end{equation}

Similarly, the CMA block facilitates interaction between the two modalities by sharing attention across them. The cross-modality refined features are computed as:

\begin{equation}
\begin{aligned}
X_{\textrm{CMA\_dm}} &= S_{\textrm{dm}} \cdot V_{\textrm{cm}} \\
X_{\textrm{CMA\_cm}} &= S_{\textrm{cm}} \cdot V_{\textrm{dm}}
\end{aligned}
\end{equation}
where $X_{\textrm{CMA\_dm}}$ represents the dermoscopic features refined using clinical information, and $X_{\textrm{CMA\_cm}}$ represents the clinical features refined using dermoscopic information. By leveraging both intra- and cross-modality attention, the model ensures that relevant information from both dermoscopic and clinical images is fully captured and integrated for a more robust melanoma diagnosis.

Next, as shown in Fig.~\ref{fig:workflow}, after the feature extraction process through the four dual-attention blocks, we obtain the refined feature embeddings $X_{\textrm{d\_refined}}$ and $X_{\textrm{c\_refined}}$ for the individual image modalities, which are combined with meta-information extracted via the SMA module. Additionally, the fusion feature $X_{\textrm{f\_refined}}$ is obtained through the CMA module. These three components are then prepared for further fusion with the information provided by the graph neural network 7PCL-DirGM.

Let $Z$ represent the features extracted from the graph information. In the next fusion step, we individually combine the graph features ($Z$) with the deep features using element-wise multiplication:

\begin{equation}
\begin{aligned}
X_{\textrm{d\_fused}} &= X_{\textrm{d\_refined}} \odot Z \\
X_{\textrm{c\_fused}} &= X_{\textrm{c\_refined}} \odot Z \\
X_{\textrm{f\_fused}} &= X_{\textrm{f\_refined}} \odot Z
\end{aligned}
\label{eq_fusion}
\end{equation}
where $X_{\textrm{d\_fused}}$, $X_{\textrm{c\_fused}}$, and $X_{\textrm{f\_fused}}$ represent the fused features for the dermoscopic, clinical, and combined channels, respectively.

Subsequently, we apply a fully connected layer to each channel independently and perform a weighted average fusion of the features:

\begin{equation}
X = \gamma_{\textrm{d}} X_{\textrm{d\_fused}} + \gamma_{\textrm{c}} X_{\textrm{c\_fused}} + \gamma_{\textrm{f}} X_{\textrm{f\_fused}}
\label{eq_weighted_fusion}
\end{equation}
where $\gamma_{\textrm{d}}$, $\gamma_{\textrm{c}}$, and $\gamma_{\textrm{f}}$ represent the weights for combining the fully connected layers of the dermoscopic, clinical, and additional channel images, respectively.

The resulting fused feature, \( X \), with dimensions \( \mathbb{R}^{n \times 768} \), is utilized for downstream tasks, where \( n \) represents the number of cases and 768 corresponds to the feature dimensionality, aligned with the meta-embedding layer. This process effectively integrates information from both the 7PCL graph and the image modalities.

\begin{algorithm}[!t]
\small
\footnotesize
\SetAlgoLined
\footnotesize 
\caption{Generation of Weighted Directed Graph}
\label{algorithm1}
\SetAlgoNlRelativeSize{0}

\KwData{List of 7PCL nodes $v_{\textrm{PCL}}$, Mel node $v_{\textrm{Mel}}$, co-directional occurrence data $\mathcal{H}$, and weighting parameters $\alpha$ and $\beta$}
\KwResult{Directed graphs $\mathcal{G}_{\textrm{internal}}$ and $\mathcal{G}_{\textrm{external}}$, and the combined graph $\mathcal{G}$}

\BlankLine
\textbf{Step 1: Generate $\mathcal{G}_{\textrm{internal}}$ - Internal Graph}

Initialize an empty directed graph $\mathcal{G}_{\textrm{internal}}$\;
Initialize a co-occurrence matrix $\mathcal{Q}$;

\For{\textnormal{each co-directional occurrence $c$ of nodes $p$ and $q$, denoted as} $(p, q, c)$ \textnormal{, in} $\mathcal{H}$}{
    Increment $\mathcal{Q}_{p\rightarrow q}$ by count $c$\;
}

\For{$p = 1$ \textnormal{to} $7$}{
    \For{$q = 1$ \textnormal{to} $7$}{
        \If{$\mathcal{Q}_{p\rightarrow q} \geq 1$}{
            Calculate the edge weight as $w_{p\rightarrow q} = \frac{\mathcal{Q}_{p\rightarrow q}}{\text{total occurrences of } v_{\textrm{PCL}_p}}$\;
            Add a directed edge from $v_{\textrm{PCL}_p}$ to $v_{\textrm{PCL}_q}$ with weight $w_{p\rightarrow q}$\;
        }
    }
}

\BlankLine
\textbf{Step 2: Generate $\mathcal{G}_{\textrm{external}}$ - External Graph}

Initialize an empty directed graph $\mathcal{G}_{\textrm{external}}$\;

\For{$p = 1$ \textnormal{to} $7$}{
    \If{$\mathcal{Q}_{p,\textrm{Mel}} \geq 1$}{
        Calculate the edge weight $w_{\textrm{Mel} \rightarrow p}$ as $w_{\textrm{Mel} \rightarrow p} = \frac{\text{total occurrences of } p}{\text{total occurrences of Mel}}$\;
        Add a directed edge from $\textrm{Mel}$ to $v_{\textrm{PCL}_p}$ with weight $w_{\textrm{Mel} \rightarrow p}$\;
    }
    \If{$\mathcal{Q}_{\textrm{Mel},p} \geq 1$}{
        Calculate the edge weight $w_{p \rightarrow \textrm{Mel}}$ as $w_{p \rightarrow \textrm{Mel}} = \frac{\text{total occurrences of Mel}}{\text{total occurrences of } p}$\;
        Add a directed edge from $v_{\textrm{PCL}_p}$ to $\textrm{Mel}$ with weight $w_{p \rightarrow \textrm{Mel}}$\;
    }
}

\BlankLine
\textbf{Step 3: Combine $\mathcal{G}_{\textrm{internal}}$ and $\mathcal{G}_{\textrm{external}}$}

Calculate $\mathcal{G} = \alpha \cdot \mathcal{G}_{\textrm{internal}} + \beta \cdot \mathcal{G}_{\textrm{external}}$\;

\BlankLine
\Return{$\mathcal{G}_{\textrm{internal}}$, $\mathcal{G}_{\textrm{external}}$, $\mathcal{G}$}
\end{algorithm}

\subsection{7-Point Checklist Directed Graph Mining (7PCL-DirGM)  }

Within the 7PCL dataset, there exist both directional and causal relationships among the various attributes, each associated with different stages of melanoma development and their causes. For instance, both irregular pigment network and atypical pigmentation are important features to consider when evaluating potential melanomas. However, irregular pigment network is often considered a more specific and reliable indicator of melanoma when compared to atypical pigmentation alone. This is because the disruption of the normal pigment network structure is a hallmark feature of melanomas and is less commonly seen in benign moles. Therefore, the presence of an irregular pigment network raises the suspicion for melanoma and may prompt further evaluation or biopsy \cite{de2018reinterpreting,pampena2020clinical,shi2020retrospective}. Recognizing these distinctions aids in filtering out superfluous and inaccurate interactions, enhancing the efficiency of graph network data utilization. Moreover, contrary to previous studies equating disease diagnosis labels with 7PCL features \cite{kawahara2018seven,bi2020multi,tang2022fusionm4net,fu2022graph}, we emphasize that the information pertaining to disease diagnosis, the various attributes, and interactions among these attributes represent significant directed information that should be considered separately with appropriate weighting for effective integration. To address these complexities, our study proposes the 7PCL-DirGM, a directional graph mining method tailored to extract vital attribute interaction data related to melanoma.

\subsubsection{Graph Node Feature Encoding}

In graph convolutional learning networks, the encoding of node information—often critical—tends to be overshadowed by the emphasis on topological relationships between nodes. Previous research frequently relies on one-hot encoding to represent textual information features; however, this approach reduces the richness of node attributes to simple labels, limiting the depth of information available for learning. For instance, positive expressions related to attributes like \textbf{pigmentation} and \textbf{streaks} often include \textbf{irregular} descriptive terms, which carry significant semantic information.

In this study, we employ the BERT embedding strategy \cite{devlin2018bert}, which not only preserves semantic context but also dynamically captures node feature details. Each node feature, denoted as $x^i_{\textrm{7p}j}$, is transformed into an embedding ${X^i_{\textrm{7p}_j} \in \mathbb{R}^{7 \times 768}}$ using the BERT module, where 768 to match the default output dimensionality of BERT. Here, 768 was chosen based on pre-experiments to optimize the representation of encoded features.

\subsubsection{Internal \& External Directional Weighted Graph (IEDWG)}
\begin{figure}[t]
    \centering
    \subfloat[External connectivity]{\includegraphics[width=0.45\columnwidth]{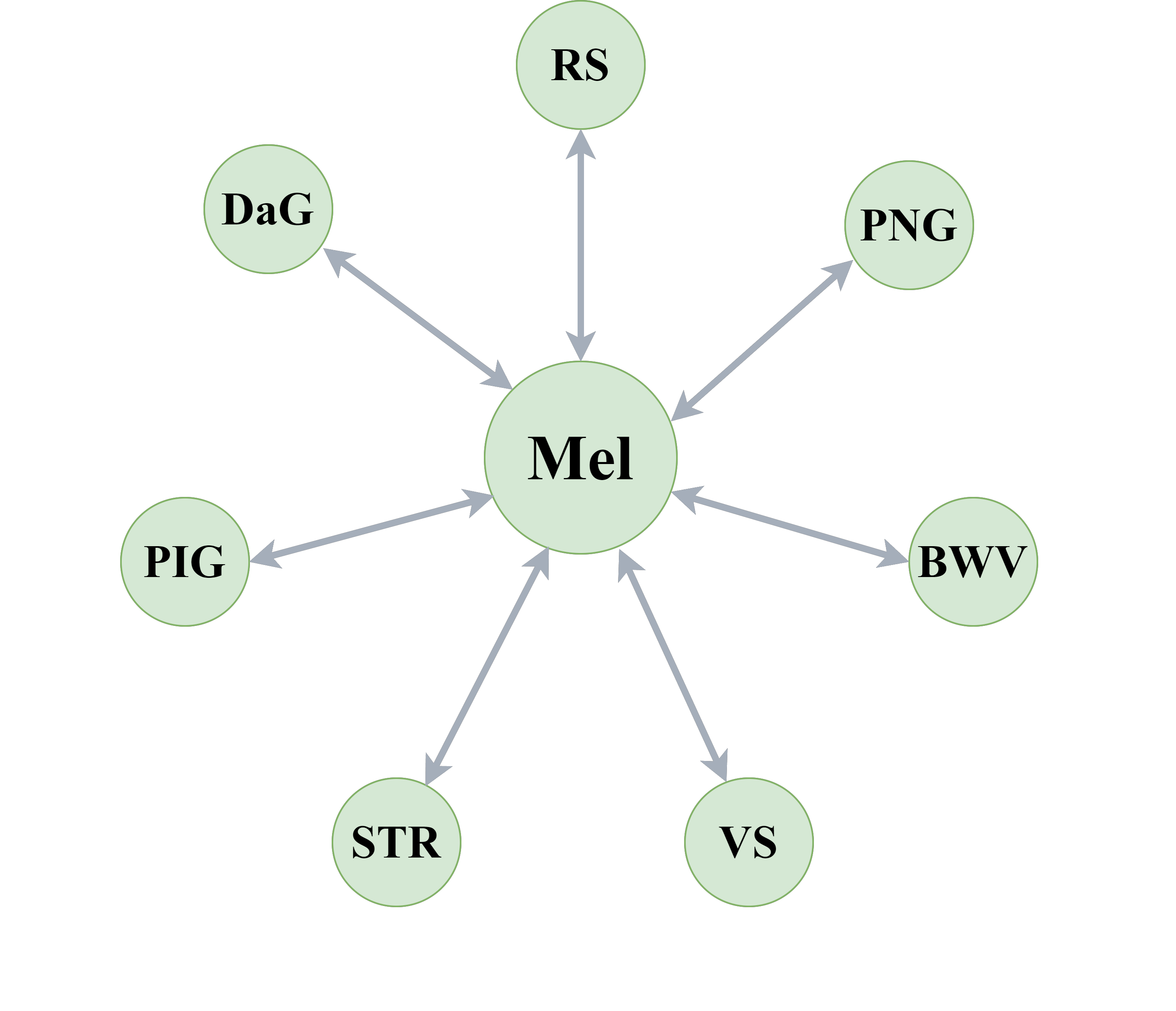}%
    \label{fig:external}}
    \subfloat[Internal connectivity]{\includegraphics[width=0.45\columnwidth]{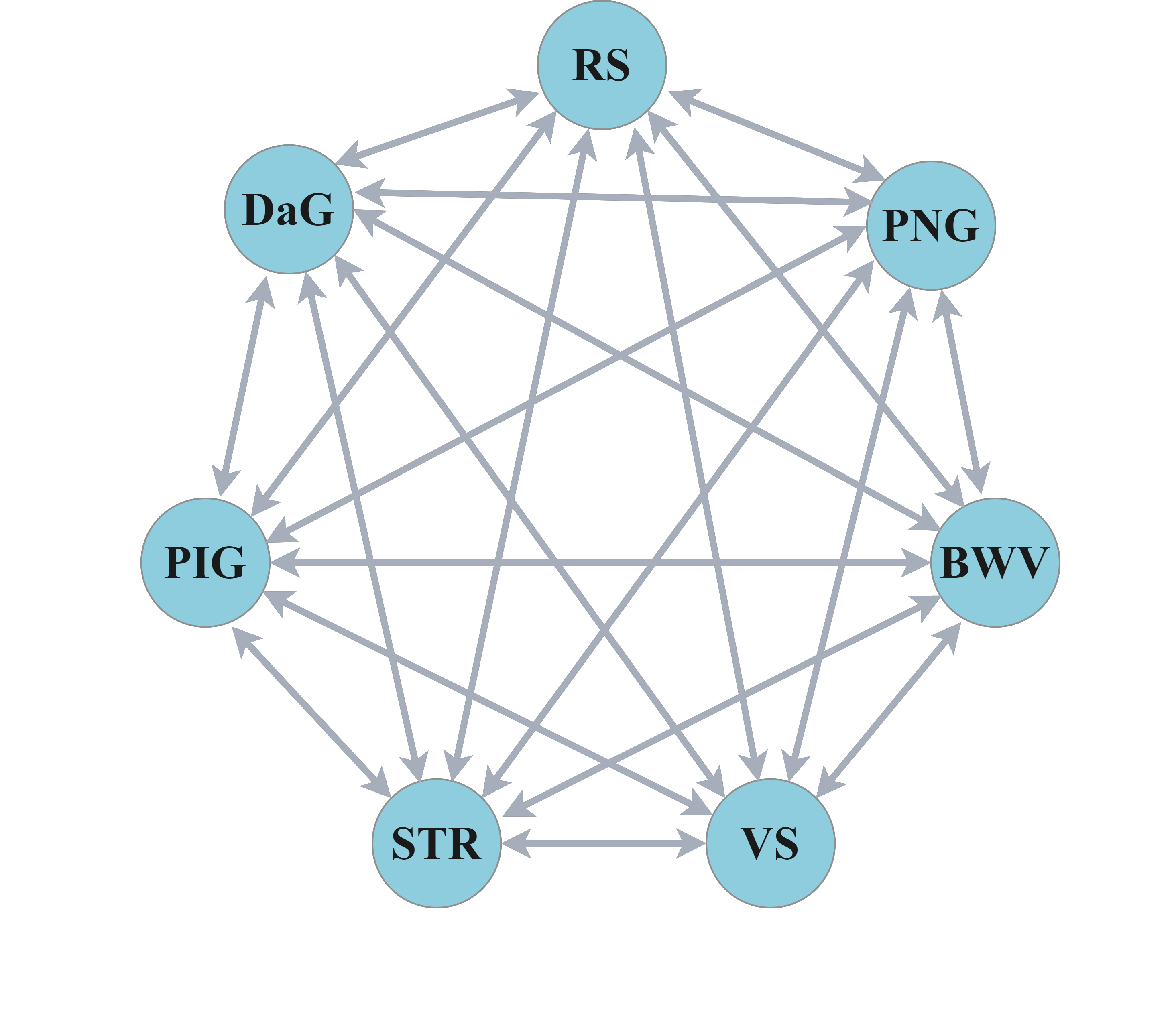}%
    \label{fig:internal}}
    
    \subfloat[Combined connectivity]{\includegraphics[width=0.9\columnwidth]{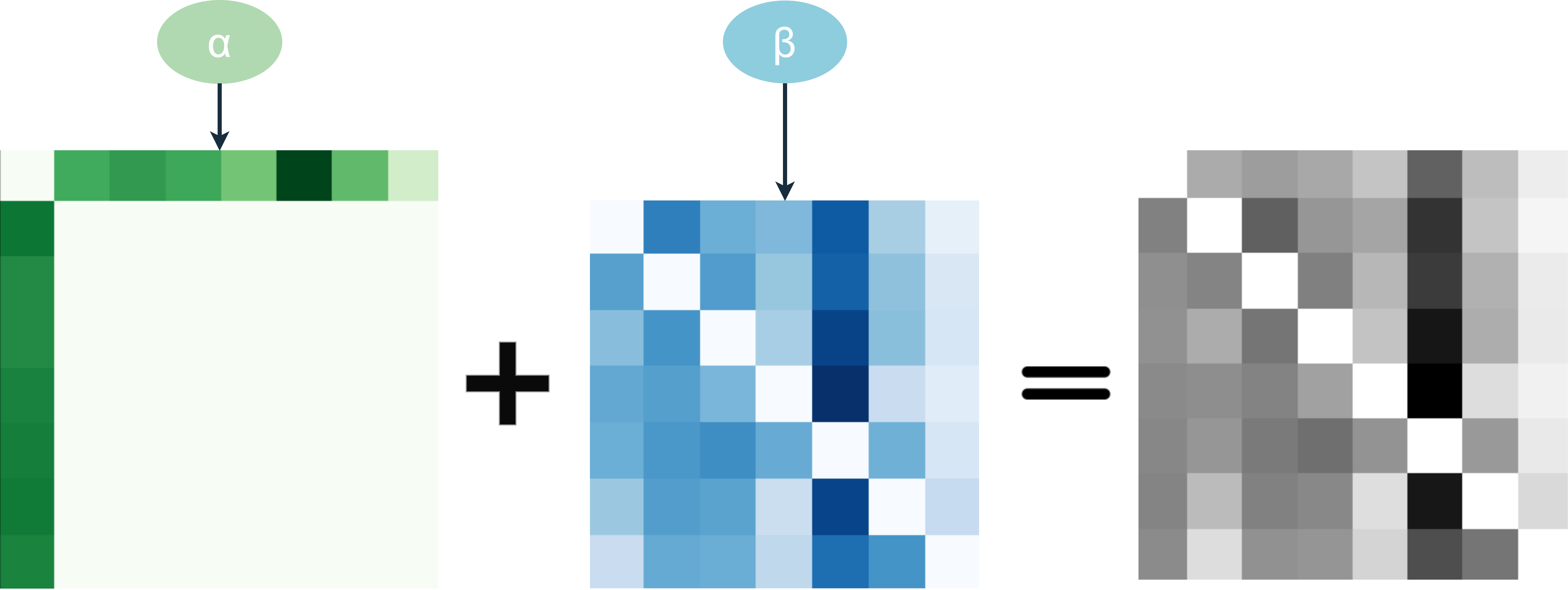}%
    \label{fig:combined}}
    \caption{The illustration depicts the principles of both internal and external connectivity. The green nodes and matrix represent external connections between the seven attributes and melanoma, while the blue nodes and matrix represent internal connections among the seven attributes themselves. The gray matrix represents the final matrix obtained by the weighted fusion of the two.}
    \label{fig: external and internal}
\end{figure}

\begin{figure}[!t]
    \subfloat[]{\includegraphics[width=0.5\columnwidth]{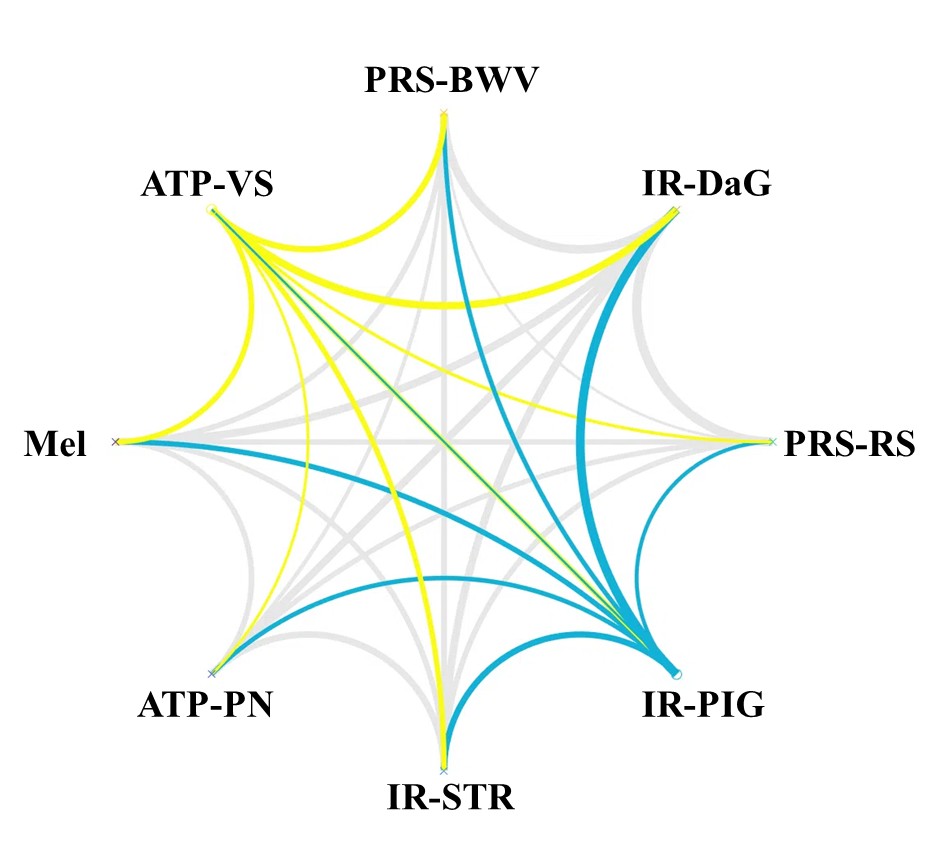}%
    \label{fig:vs_pig_graph}}
    \subfloat[]{\includegraphics[width=0.5\columnwidth]{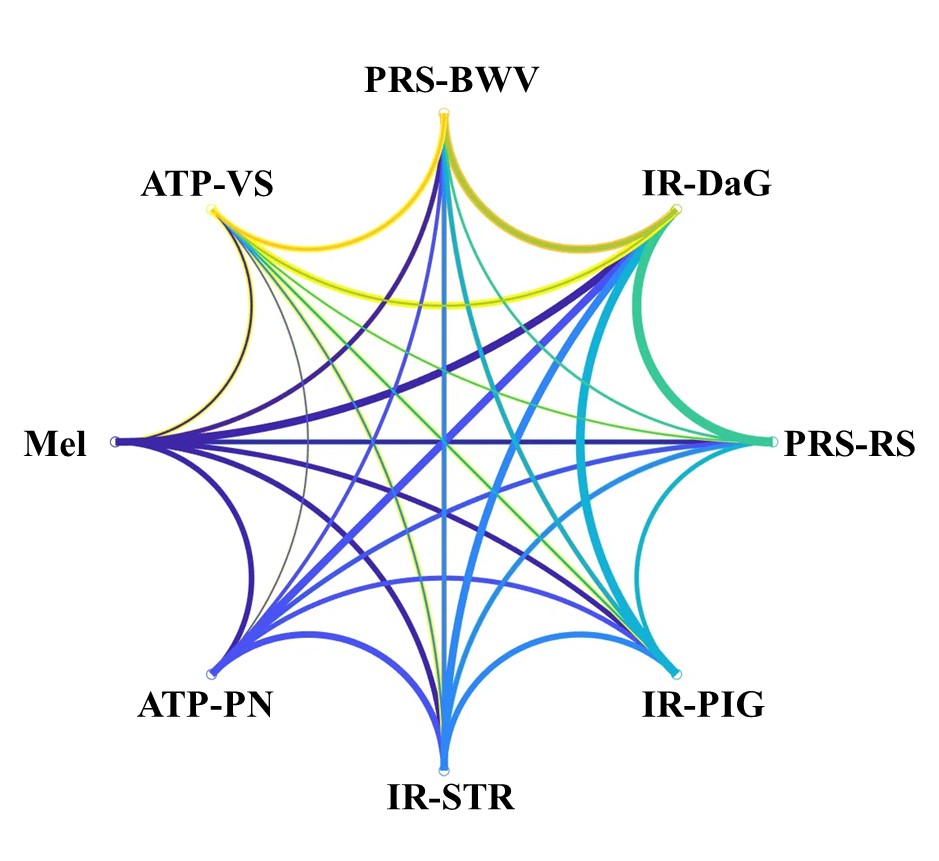}%
    \label{fig:all_graph}}
    
    \caption{The plot illustrates the weighted connections in the training set of the EDRA dataset. while (a) focuses on a specific pair, ATP-VS and IR-PIG. Yellow lines represent connections from ATP-VS to the other nodes, while cyan lines represent connections from IR-PIG. Variations in line thickness highlight differences in directed weights between the nodes. Figure (b) displays the overall connectivity map among all eight nodes.}
    \label{fig:directed connection}
\end{figure}

One significant deviation from related studies is the importance we place on the directed connection between attributes in 7PCL for constructing directed graph networks. As introduced in \textbf {Algorithm} \ref{algorithm1} and Fig.~\ref{fig: external and internal}, this connection is reflected primarily in two levels: the internal level within the attributes, which is determined by the mutual connections within the seven attributes in the checklist. The external level involves the coexistence probability of each attribute and the melanoma. Both internal and external information are used to build the weighted directional edges by using the conditional coexistence probability directedness $w_{p,q}$, which is calculated based on the training data. This probability serves as the weight of all edges and illustrates the unequal dependency relationship among attributes, as shown in Fig.~\ref{fig:directed connection}.

\subsubsection{Adaptive Receptive Field Proximity (ARFP)}
Accurately modeling attribute relationships is crucial for melanoma diagnosis using the 7PCL. However, conventional graph-based methods struggle to capture long-range dependencies, leading to imbalanced receptive fields and suboptimal feature learning. To address this, we introduce ARFP, which enhances the multi-order connectivity of the 7PCL-directed weighted graph, improving melanoma classification. Inspired by digraph inception convolutional networks~\cite{tong2020digraph}, ARFP leverages the $k^\text{th}$-order proximity to incorporate both direct and indirect attribute interactions. Given a graph $\mathcal{G} = (\mathcal{V}, \mathcal{E})$, where $\mathcal{V}$ represents 7PCL attributes and $\mathcal{E}$ represents their directed relationships, the proximity paths are defined as: 

\begin{equation}
\label{path}
\left\{
\begin{aligned}
\mathcal{M}_{p, q}^{(k)} &= v_{p} \underbrace{\rightarrow \cdots \rightarrow}_{k-1 \text { edges }} v_{e} \underbrace{\leftarrow \cdots \leftarrow}_{k-1 \text { edges }} v_{q}, \\
\mathcal{D}_{p, q}^{(k)} &= v_{p} \underbrace{\leftarrow \cdots \leftarrow }_{k-1 \text { edges }} v_{e} \underbrace{\rightarrow \cdots \rightarrow}_{k-1 \text { edges }} v_{q}.
\end{aligned}
\right.
\end{equation}
where $\mathcal{M}_{p, q}^{(k)}$ and $\mathcal{D}_{p, q}^{(k)}$ represent the meeting and diffusion paths between attributes $p$ and $q$ $(p, q \in \mathcal{V})$. If both paths exist, the attributes are at the $k^\text{th}$-order proximity level, with $v_{e}$ as their common neighbor.

Using the 7PCL-directed weighted graph and the $k^\text{th}$-order proximity, the 7PCL $k^\text{th}$-order proximity matrix founded on the conventional spectral-based GCN method is denoted as:

\begin{equation}
\label{kth-order proximity}
\resizebox{.9\columnwidth}{!}{$
\begin{aligned}
{P}^{(k)} = \begin{cases}
    \multicolumn{1}{c}{{I}} &  k=0 \\
    \multicolumn{1}{c}{{{D}}^{-1} {{A}}} &  k=1 \\
    \multicolumn{1}{c}{\operatorname{Intersect}\left(\left({P}^{(1)}\right)^{k-1}\left({P}^{(1)^T}\right)^{k-1},\left({P}^{(1)^T}\right)^{k-1}\left({P}^{(1)}\right)^{k-1}\right) / 2} &  k \geq 2
\end{cases}
\end{aligned}
$}
\end{equation}
where ${{A}}$ represents the adjacency matrix of the directed weighted graph ${\mathcal{G}}$, and ${{D}}$ represents the corresponding diagonalized degree matrix. The matrix intersection operation $\text{intersect}(\cdot)$ creates a new matrix with the element-wise intersection of both meeting and diffusion paths. This operation aims to symmetrize the $k^{th}$-order proximity matrix ${P}^{(k)}$ to fit the structure of spectral-based GCN. Here, the value of parameter $k$ represents the distance between two attributes, which determines the size of the receptive fields. By adjusting the value of parameter $k$, users can observe and select the optimal receptive field with scalable capabilities.

Based on the $k^{th}$-order proximity matrix, the multi-scale digraph is then defined as: 

\begin{equation}
\label{adaptive Z}
\resizebox{.9\columnwidth}{!}{$
\begin{aligned}
{Z}^{(k)}=\left\{\begin{array}{cc}
{X} \Theta^{(0)} & k=0 \\
\frac{1}{2}\left(\boldsymbol{\Pi}^{(1)^{\frac{1}{2}}} {P}^{(1)} \boldsymbol{\Pi}^{(1)^{-\frac{1}{2}}}+\boldsymbol{\Pi}^{(1)^{-\frac{1}{2}}} {P}^{(1)^T} \boldsymbol{\Pi}^{(1)^{\frac{1}{2}}}\right) {X} \Theta^{(1)} & k=1 \\
{W}^{(k)^{-\frac{1}{2}}} {P}^{(k)} {W}^{(k)^{-\frac{1}{2}}} {X} \Theta^{(k)} & k \geqslant 2
\end{array} \right.
\end{aligned}
$}
\end{equation}
where ${Z}^{(k)}$ represents the convolved output with dimensions $\mathbb{R}^{n \times d}$. ${X}$ denotes the node feature matrix with dimensions $\mathbb{R}^{n \times c}$. ${W}^{(k)}$ refers to the weight matrix, which is diagonalized from ${P}^{(k)}$, and ${\Theta}^{(k)}$, with dimensions $\mathbb{R}^{c \times d}$, is a trainable weight matrix. It's noteworthy that when $k=1$, ${Z}^{(1)}$ is computed through digraph convolution using ${P}^{(1)}$ as defined in Eq.~\ref{kth-order proximity}, and $\boldsymbol{\Pi}^{(1)}$ represents the corresponding approximate diagonalized eigenvector associated with ${P}^{(1)}$.

This approach enables a scalable representation of 7PCL attribute relationships, ensuring that clinically meaningful dependencies are preserved and effectively utilized in melanoma diagnosis. By leveraging the adaptive receptive field, our framework refines multi-scale learning across directed attributes, improving classification performance and interpretability.  

\subsection{Gradient Diagnostics with Data-Driven Weighting (GD-DDW)}

As previously discussed, the deep features extracted from fused imaging modalities and the directed mutual connection matrix are collectively referred to as \(X\) for subsequent diagnostic procedures. In this section, we present the GD-DDW method, designed to emulate the diagnostic approach employed by dermatologists using the 7-point checklist algorithm in clinical practice. Specifically, the GD-DDW method utilizes representative features corresponding to various attributes to make initial predictions, which are then used to diagnose melanoma. 

In this module, we first establish a parallel multi-label classifier designed to address the classification of seven attributes on the checklist. As \(X\) is the input, \(y_{{\textrm{7p}}_j}= [y_{{\textrm{7p}}_1}, y_{{\textrm{7p}}_2}, ..., y_{{\textrm{7p}}_7}] \in \{0, 1\}^7\) is denoted as the ground truth of all seven attributes in the format of a binary indicator vector. In the proposed method, the focal loss function \cite{lin2017focal} was employed with the objective of enhancing the performance on the dataset that exhibiting a serious imbalance.

The focal loss function for each label can be defined as:
\begin{equation}
L_{{\textrm{7p}}_j} = -\mu_j (1 - \hat{y}_{{\textrm{7p}}_j})^\tau \log(\hat{y}_{{\textrm{7p}}_j})
\label{focal_loss}
\end{equation}
where \(\hat{y}_{{\textrm{7p}}_j}\) represents the predicted probability of the positive class, \(\mu_j\) is a balancing parameter, and \(\tau\) is the focusing parameter. The total focal loss \(L_{\textrm{7p}}\) for all seven attributes can be formulated as:

\begin{equation}
L_{\textrm{7p}} = \sum_{j=1}^{7} L_{{\textrm{7p}}_j}
\label{focal_loss_7pt}
\end{equation}

Next, the prediction scores obtained for each label are utilized as inputs for the subsequent step. In this stage, a weighted sum function and a sigmoid activation function are applied to perform classification for the diagnosis of melanoma. 

Let \(w = [w_1, w_2, ..., w_j]\) represent the learned weights module obtained from the attributes. In this module, the predicted probability of melanoma \(\hat{y}_\textrm{m}\) is computed as:

\begin{equation}
\hat{y}_\textrm{m} = \sigma\left(\frac{1}{\sum_{j=1}^{7} w_j} \sum_{i=j}^{7} w_j \hat{y}_{{\textrm{7p}}_j}\right)
\label{prediction_mel}
\end{equation}
where \(\sigma(\cdot)\) denotes the sigmoid activation function and the rescaling factor \(\frac{1}{\sum_{j=1}^{7} w_j}\) ensures that the output is appropriately scaled to the range \([0, 1]\).

The overall loss \(L\) is then computed as the sum of \(L_{\textrm{7p}}\) and \(L_{\textrm{mel}}\):

\begin{equation}
L = { L_{\textrm{7p}}} + {\lambda L_{\textrm{mel}}}
\label{overall_loss}
\end{equation}
where $\lambda$ is the weighting parameter that adjusts the contributions of the two losses, specifically, making sure the loss values are in the same order of magnitude.

\section{Experiments}

\subsection{Materials}
In this study, we utilized the publicly available EDRA dataset (also known as Derm7pt), specifically curated for 7PCL studies and annotated by Kawahara et al. \cite{kawahara2018seven}. This dataset comprises paired dermoscopic and clinical images sourced from 1011 patients, each image has a maximum resolution of 768 by 512 pixels. Notably, nine patients lacked clinical images, which were substituted with corresponding dermoscopic ones. Throughout our study, we strictly adhered to the dataset's official specifications, which stipulated a total of 413 training, 203 validation, and 395 test images. Concerning the labels, the 7PCL attributes were incorporated into the analysis, following the official categorization. It is worth noting that the original disease classification (DIAG) system included specific disease labels for MISC, BCC, Mel, SK, and Nevi. This study primarily focused on distinguishing melanoma from lookalike diseases. To verify the generalizability of our proposed method, we also used two publicly available datasets, ISIC 2017 and ISIC 2018 \cite{codella2018skin}, which also have multiple attributes. The visual attributes contained in ISIC 2017 include milia-like cysts, negative network, pigment network, and streaks. A total of 2000 training, 150 validation, and 600 test images. ISIC 2018 adds an additional attribute, globules, to the 2017 version and adds a portion of the data to make the data distribution adjusted to 2594 training, 100 validation, and 1000 test images.

\subsection{Implementation Details}

In our study, we ensured the integrity of our results by employing consistent data processing methods and adhering to uniform model training protocols across various ablation comparison experiments. Our chosen deep learning model framework is based on three residual blocks. The entire architecture is meticulously fine-tuned end-to-end, utilizing the Adam optimizer \cite{zhang2018improved} with a concurrent learning rate set to 0.00001. We predefined the number of epochs for pretraining to 150, with an early-stopping mechanism in place. If the model's performance on the validation set fails to improve over 50 consecutive epochs, the training process halts, and the best-performing model is saved. The loss functions utilized are based on the focal loss to circumvent the result bias that may arise from imbalanced data. The hyperparameters for the different stages outlined in the methodology were selected based on insights gained from the training process and represent approximate ranges. The initial stage of the process is the pre-processing of the images. To ensure consistency and minimize the impact of varying image formats, we removed black edges from the images and resized them to 512 by 512 pixels, maximizing the use of image features while avoiding information loss.

During the data preparation phase, all labels are transformed into standardized textual descriptions. These are then processed by the semantic embedding function to effectively utilize label information. Each patient case follows the format: " This patient has been diagnosed with [Diagnosis], exhibiting the attributes [Attribute 1], [Attribute 2], and [Attribute 3]. " This structured representation ensures consistency in data processing and facilitates the meaningful integration of clinical attributes into the model. The standard texts of the indicated 7PCL are atypical pigment network (ATP-PN), irregular streaks (IR-STR), irregular pigmentation (IR-PIG), regression structure (RS), irregular dots and globus (IR-DaG), blue-whitish veil (BWV), and irregular vascular structures (IR-VS).

In the CD-MFM section, features from different channels undergo two stages of fusion using weighted averaging. Initially, a fusion process occurs between clinical and dermoscopy images using the dual-attention blocks, followed by a second fusion stage after incorporating topological relations learned by the graph network. The parameters $\gamma$ series in Eq.~\ref{eq_weighted_fusion} are configured to ${1}/{3}$ to mitigate potential biases stemming from assigning varying weights to the same feature across multiple instances. In constructing the graph network, we consider only a subset of the possible connections between nodes due to the directionality of the multistep connections. Based on the average number of co-existing labels in the training set and the avoidance of isolated points, the connections are ordered according to the edge weights, ensuring that each node has at least one and no more than three edges. The order of connection $k$ is set to 3 due to the natural limitation of the number of nodes in 7PCL. In the final diagnostic phase, the parameter $\lambda$, utilized in the global optimization of the loss function in Eq.~\ref{overall_loss}, is set to 1, given that both loss values are of a comparable magnitude. All experiments were conducted on one NVIDIA Tesla V100 GPU card (32 GB memory). The duration of each training cycle is approximately one hour. 

\subsection{Evaluation Settings}

The study used standard performance metrics such as AUC, sensitivity and specificity. To demonstrate our method's advantages, we compared it with state-of-the-art methods like RemixFormer++ \cite{xu2024remixformer++}, CAFNet \cite{he2023co}, GRM-DC \cite{fu2022graph}, 7PCL-Constrained-CC \cite{wang2021incorporating}, HcCNN \cite{bi2020multi}, TripleNet \cite{ge2017skin}, CCRD \cite{wang2023consistent}, GELN \cite{tang2023graph}, and EmbeddingNet \cite{yap2018multimodal}. Specifically, using the Inception models introduced by Kawahara et al. \cite{kawahara2018seven} as the baseline for comparison, as they originally published the data along with the method.

We also conducted ablation experiments to evaluate the critical components of our method. These experiments included comparisons between unimodal and multimodal fusion (Dual-Attention), assessments of our directional graph mining-based GCN strategy (7PCL-DirGM), and tests of our gradient-driven prediction structure (GD-DDW) against parallel architectures.

In addition, the prediction results for all attributes obtained from the first layer of the GD-DDW structure were utilized. Following the traditional 7PCL rule, diagnostic scores were calculated by summing the assigned attribute scores. Thresholds ranging from 1 to 7 were applied to diagnose melanoma. These results were then compared with the proposed method across two tasks: Mel vs. Nevi and Mel vs. other diseases, including MISC, SK, Nevi, and BCC.

We further validated our method using two additional publicly available datasets, ISIC 2017 and ISIC 2018. Although these datasets provide only attribute information and lack diagnostic labels, preventing the evaluation of GD-DDW's impact, they still allow the validation of directed multi-order relationships between the attributes (CKTG).

\begin{table*}[!t]
\scriptsize
\setlength{\tabcolsep}{3pt}
\caption{Classification Performance (AUC, Sens, and Spec) of the Proposed Method Compared with the State-of-the-Art Methods, Evaluated on the EDRA Dataset. Top two AUC scores per label are highlighted.}
\centering
\label{tab:my-table-sota}
\def\tblwidth{\textwidth}
\begin{tabular*}{\tblwidth}{@{\extracolsep{\fill}} l >{\raggedright\arraybackslash}p{3cm} *{9}{l}}
\toprule
 &
  \textbf{Method} &
  \textbf{\begin{tabular}[c]{@{}c@{}}DIAG\\ MEL\end{tabular}} &
  \textbf{\begin{tabular}[c]{@{}c@{}}PN\\ ATP\end{tabular}} &
  \textbf{\begin{tabular}[c]{@{}c@{}}STR\\ IR\end{tabular}} &
  \textbf{\begin{tabular}[c]{@{}c@{}}PIG\\ IR\end{tabular}} &
  \textbf{\begin{tabular}[c]{@{}c@{}}RS\\ PRS\end{tabular}} &
  \textbf{\begin{tabular}[c]{@{}c@{}}DaG\\ IR\end{tabular}} &
  \textbf{\begin{tabular}[c]{@{}c@{}}BWV\\ PRS\end{tabular}} &
  \textbf{\begin{tabular}[c]{@{}c@{}}VS\\ IR\end{tabular}} &
  \textbf{Avg} \\ \midrule

                                \multirow{8}{*}{\textbf{AUC}} 
                                 & \textbf{Inception-combined \cite{kawahara2018seven}}   & 86.3 & 79.9 & 78.9 & 79.0 & 82.9 & 79.9 & 89.2 & 76.1 & 81.5 \\
                                 & \textbf{EmbeddingNet \cite{yap2018multimodal}}         & 82.5 & 74.5 & 77.7 & 77.9 & 71.3 & 78.5 & 84.8 & 76.9 & 78.0 \\
                                 & \textbf{TripleNet \cite{ge2017skin}}                   & 81.2 & 73.8 & 76.3 & 77.6 & 76.8 & 76.0 & 85.1 & 79.9 & 78.3 \\
                                 & \textbf{Constrained-CC \cite{wang2021incorporating}}   & 83.4 & 76.2  & 81.8 & 82.6 & 78.4 & 81.1 & 90.5 & 78.6 & 81.6 \\
                                 & \textbf{HcCNN \cite{bi2020multi}}                      & 85.6 & 78.3 & 77.6 & 81.3 & 81.9 & 82.6 & 89.8 & 82.7 & 82.5 \\
                                 & \textbf{GRM-DC \cite{fu2022graph}}                     & 87.6 & \textbf{87.5} & 81.2 & 83.6 & 79.0 & 83.1 & 90.8 & 75.4 & 83.5 \\
                                 & \textbf{CAFNet \cite{he2023co}}                        & \textbf{92.2} & 75.3 & \textbf{85.4} & \textbf{85.0} & 85.4 & 78.7 & \textbf{94.7} & 83.4 & 84.5 \\
                                 & \textbf{CCRD \cite{wang2023consistent}}              
                                    &   ~\textemdash 
                                    & 82.5  
                                    & 82.8  
                                    & 82.1
                                    & 84.8   
                                    & 83.0
                                    & 92.5
                                    & 84.1
                                    & 84.5 \\
                                    
                                    & \textbf{GELN \cite{tang2023graph}}             
                                    & 90.2 
                                    & 84.0 
                                    & 82.3  
                                    & 83.8
                                    & 82.9   
                                    & 81.9
                                    & 91.8 
                                    & \textbf{84.5}
                                    & 85.2  \\

                                 & \textbf{RemixFormer++ \cite{xu2024remixformer++}}      & 91.3 & \textbf{87.7} & 81.6 & 82.5 & \textbf{86.1}& \textbf{84.2} & 91.1 & \textbf{86.2} & \textbf{86.3} \\
                                 & \textbf{Ours}                                           & \textbf{92.0}& 86.7 & \textbf{90.1} & \textbf{88.3} & \textbf{89.3} & \textbf{86.2} & \textbf{93.4} & 83.0 & \textbf{88.6} \\ \midrule

\multirow{8}{*}{\textbf{Sens}}
                                 & \textbf{Inception-combined \cite{kawahara2018seven}}   & 61.4 & 48.4 & 51.1 & 59.7 & 66.0 & 62.1 & 77.3 & 13.3 & 54.9 \\
                                 & \textbf{EmbeddingNet \cite{yap2018multimodal}}         & 40.6 & 33.3 & 51.1 & 60.5 & 96.2 & 64.4 & 96.3 & 23.3 & 58.2 \\
                                 & \textbf{TripleNet \cite{ge2017skin}}            & 46.5 & 33.3 & 39.4 & 61.3 & 97.9 & 67.2 & 90.0 & 30.0 & 58.2 \\
                                 & \textbf{Constrained-CC \cite{wang2021incorporating}}   & 65.3  & 39.8  & 52.1  & 58.5  & 40.6 & 63.8  & 73.3  & 30.0  & 52.9 \\
                                 & \textbf{HcCNN \cite{bi2020multi}}                & 58.4 & 40.9 & 35.1 & 55.7 & 95.2 & 80.2 & 92.2 & 20.0 & 59.7 \\
                                 & \textbf{GRM-DC \cite{fu2022graph}}               & 59.0 & 77.5 & 67.0 & 39.2 & 21.9 & 70.1 & 69.9 & 3.6  & 51.0 \\

                                & \textbf{CAFNet \cite{he2023co}}              & 75.3  & 65.9  & 67.1 & 60.3 & 42.7 & 74.1 & 68.8 & 45.0 & 62.4 \\
                                & \textbf{CCRD \cite{wang2023consistent}}      & ~\textemdash & 67.8 & 63.2 & 69.0 & 80.0 & 75.0 & 87.5 & 100.0 & 77.5 \\
                                & \textbf{GELN \cite{tang2023graph}}           & 68.1 & 57.4 & 57.0 & 68.9 & 74.6 & 70.0 & 76.7 & 50.0 & 65.3 \\

                                 & \textbf{RemixFormer++ \cite{xu2024remixformer++}}               & 72.3 & 61.3 & 67.0 & 63.7 & 47.2 & 76.8 & 72.0 & 33.3 & 61.7 \\
                                 
                                                                  & \textbf{Ours}  & 74.3 & 63.4 & 76.6 & 56.5 & 79.2 & 85.9 & 82.7 & 40.0 & 69.8 \\ \midrule

\multirow{8}{*}{\textbf{Spec}}
                                 & \textbf{Inception-combined \cite{kawahara2018seven}}   & 88.8 & 90.7 & 85.7 & 80.1 & 81.3 & 78.9 & 89.4 & 97.5 & 86.6 \\
                                 & \textbf{EmbeddingNet \cite{yap2018multimodal}}         & 93.5 & 90.7 & 88.7 & 82.7 & 20.8 & 78.4 & 52.0 & 96.7 & 75.4 \\
                                 & \textbf{TripleNet \cite{ge2017skin}}            & 90.1 & 93.0 & 92.4 & 76.8 & 23.6 & 71.6 & 60.0 & 98.6 & 75.8 \\
                                 & \textbf{Constrained-CC \cite{wang2021incorporating}}   & 81.6  & 93.7  & 90.4  & 75.7 & 95.6  & 85.3 & 91.6 & 99.2  & 89.1  \\
                                 & \textbf{HcCNN \cite{bi2020multi}}                & 88.1 & 92.4 & 90.0 & 86.3 & 41.5 & 71.6 & 65.3 & 98.4 & 79.2 \\
                                 & \textbf{GRM-DC \cite{fu2022graph}}               & 89.5 & 79.0 & 80.3 & 95.8 & 96.8 & 78.8 & 91.0 & 100.0  & 88.9 \\                            
                                 & \textbf{CAFNet \cite{he2023co}}              & 93.6 & 90.2 & 91.2 & 89.1 & 96.5 & 74.0 &95.1 & 98.7 &91.1\\
                                & \textbf{CCRD \cite{wang2023consistent}}                
                                & ~\textemdash 
                                & 85.4  
                                & 85.6 
                                & 79.2 
                                & 81.8 
                                & 74.1 
                                & 90.5  
                                & 0.0
                                & 70.9  \\ 
                                
                                & \textbf{GELN \cite{tang2023graph}}                 
                                & 87.7 
                                & 83.5 
                                & 84.5 
                                & 81.9 
                                & 81.6 
                                & 76.8  
                                & 91.3
                                & 92.6
                                & 85.0  \\

                                 & \textbf{RemixFormer++ \cite{xu2024remixformer++}}               & 93.9 & 89.7  & 79.7 & 84.5 & 94.8 & 75.2 & 89.7 & 97.3 & 88.1   \\
                                                                  & \textbf{Ours}  & 90.8 & 89.1 & 85.4 & 87.5 & 81.7 & 65.6 & 89.1 & 95.8 & 85.6 \\\bottomrule

\end{tabular*}
\end{table*}

\section{Results}

\subsection{Comparison to State-of-the-Art Methods}

Table~\ref{tab:my-table-sota} presents a detailed comparison of the classification performance between our proposed method and state-of-the-art deep learning-based techniques on the EDRA dataset. In this table, the top two AUC values for each label have been bolded to emphasize the highest-performing methods. Our approach achieved an average AUC of 88.6\%, demonstrating strong generalization and robustness across multiple diagnostic labels. Specifically, it attained the highest AUC for four of the eight diagnostic labels: STR (90.1\%), PIG (88.3\%), RS (89.3\%), and DaG (86.2\%). For the remaining labels, it consistently ranked among the top two, underscoring its reliability across different classification tasks.

Building on this, when analyzing the comparative performance of existing models, all methods evaluated in Table~\ref{tab:my-table-sota} are deep learning-based. However, only GRM-DC and GELN incorporate GCNs, which explicitly model relationships between clinical attributes. Our approach further advances this direction by introducing the consideration of directed interactions between attributes and melanoma, leading to improved experimental results. These findings indicate that modeling directed relationships enhances diagnostic performance and contributes to a more interpretable model.

As the most recent state-of-the-art (SOTA) method aside from our proposed approach, RemixFormer++ integrates a transformer-based feature extractor with clinical diagnostic logic, achieving a strong performance. In contrast, our method further embeds clinical rules into the graph network while incorporating the self-attention mechanism used in transformers to enable multimodal task learning. This design reduces model size while enhancing performance. In the eight-label classification task, our approach achieved improvements in six categories, increasing the average AUC from 86.3\% to 88.6\%.

Although our proposed method achieves improvements in both AUC and sensitivity, its advantage over other methods in specificity (85.6\%) is not as pronounced. This is primarily due to the detection mechanism of the 7PCL clinical algorithm, which is designed to maximize the identification of potential melanoma cases and ensure that positive patients are not overlooked. As a result, it inherently favors high sensitivity over specificity  \cite{elwood2018skin}. In our proposed method, we address this issue by optimizing the weight distribution and threshold selection, enabling the model to maintain a high sensitivity (69.8\%) while achieving competitive specificity. This balanced approach enhances the reliability of the results and makes the model more clinically applicable.

\subsection{Ablation Studies}

The proposed methodological framework consists of multiple modules, each serving a distinct role: CD-MFM facilitates comprehensive multimodal feature learning, 7PCL-DirGM enhances graph-based representation learning with directed interactions, and GD-DDW integrates clinical diagnostic knowledge to improve interpretability and decision-making. To systematically evaluate the contribution of each module, Table~\ref{tab:my-table-ablation} presents a comparative analysis through an ablation study.

The first comparison assesses unimodal inputs, where dermoscopic and clinical images are processed separately, against multimodal fusion using dual-attention blocks. The results indicate that integrating dermoscopic and clinical imaging via dual-attention mechanisms improves the average AUC by 7\% to 10\%, highlighting the benefits of leveraging complementary information from different imaging modalities. Additionally, dermoscopic images alone outperform clinical images by an average of 3\%, which aligns with expectations since dermoscopy provides enhanced feature visibility, particularly in the context of attribute-based analysis within the 7PCL framework.

Further analysis examines the individual impact of the 7PCL-DirGM and GD-DDW modules. When incorporated independently, both modules contribute to an average AUC improvement of 9\% over the baseline dermoscopic model, with a particularly notable 13\% increase in melanoma classification. These findings emphasize the critical role of incorporating structured graph-based relationships and gradient-based diagnostic decision weighting, both of which enhance the model’s ability to capture clinically relevant feature interactions. 

The final evaluation considers the combined integration of all three modules, demonstrating a synergistic effect that further elevates model performance. The full framework achieves an average AUC of 87\%, representing a significant enhancement over individual components. These results validate the effectiveness of combining multimodal feature learning, directed graph modeling, and diagnostic decision weighting, reinforcing the importance of clinically guided deep learning approaches for melanoma diagnosis.

\begin{table*}[t]
\footnotesize
\centering
\caption{Comparative Ablation Study of Key Modules in Our Proposed Method, Evaluated on AUC using the EDRA Dataset.}
\label{tab:my-table-ablation}
\begin{adjustbox}{width=0.9\textwidth}
\begin{tabular}{cc|cccccccc}
\hline
\multirow{2}{*}{\textbf{}} &
\multirow{2}{*}{\textbf{}} &
\multirow{2}{*}{\textbf{Baseline (Clinical) }}&
  \multirow{2}{*}{\textbf{Baseline (Derm)}}&\multirow{2}{*}{\textbf{
  +Dual-Attention}}&\multirow{2}{*}{\textbf{+7PCL-DirGM}} &
\multirow{2}{*}{\textbf{+GD-DDW}} &\multirow{2}{*}{\textbf{Ours}}\\
\textbf{} & & &   &&& &  &\\
\hline
\multirow{5}{*}
\textbf{DIAG}
& \textbf{MEL} &75.6 &78.2 & 89.6 & 91.5 & 91.2 & 92.0 \\
\textbf{PN}
& \textbf{TYP} & 71.7 & 75.8 & 72.4 & 77.6 & 79.8 & 84.6  \\
& \textbf{ATP} & 69.5 & 66.9 & 76.2 & 78.0 & 77.4 & 86.7   \\
\textbf{STR}
& \textbf{REG} & 79.2 & 84.2 & 79.1 & 81.2 & 82.7 & 88.1 \\
& \textbf{IR}  & 58.5 & 67.9 & 80.2 & 81.7 & 80.6 & 90.1  \\
\textbf{PIG}
& \textbf{REG} & 66.7 & 67.8 & 81.5 & 80.8 & 82.7 & 86.1  \\
& \textbf{IR}  & 68.1 & 72.0 & 81.2 & 81.0 & 81.1 & 88.3  \\
\textbf{RS}
& \textbf{PRS} &58.4 &72.9 & 76.2 & 78.6 & 78.9 &89.3  \\
\textbf{DaG}
& \textbf{REG} & 69.6 & 67.7 & 80.0 & 81.8 & 82.2 & 84.5  \\
& \textbf{IR}  & 67.1 & 71.3 & 80.4 & 82.0 & 80.5 & 86.2  \\
\textbf{BWV}
& \textbf{PRS} &79.6 & 83.6 & 92.3 & 93.3 & 92.3 &93.4  \\
\textbf{VS}
& \textbf{REG} &71.5 & 75.6 & 72.5 & 75.5 & 76.3 & 78.9 \\
& \textbf{IR} & 66.4 & 64.3 & 73.6 & 76.6 & 75.8 & 83.0\textbf{ }\\
\textbf{Avg}
& & 69.4&  72.9&79.6&81.5 &81.7  & 87.0  \\
\hline
\end{tabular}
\end{adjustbox}
\end{table*}

\begin{table}[b]
\footnotesize
\caption{Comparison of Weights Value between Proposed Method and Traditional Algorithm under Different Tasks.}
\label{tab:my-table-weights}
\resizebox{\columnwidth}{!}{%
\begin{tabular}{llccccccc} 
\hline
\textbf{Task}    &                      & \textbf{\begin{tabular}[c]{@{}c@{}}PN\\ ATP\end{tabular}} & \textbf{\begin{tabular}[c]{@{}c@{}}STR\\ IR\end{tabular}} & \textbf{\begin{tabular}[c]{@{}l@{}}PIG\\ IR\end{tabular}} & \textbf{\begin{tabular}[c]{@{}c@{}}RS\\ PRS\end{tabular}} & \textbf{\begin{tabular}[c]{@{}c@{}}DaG\\ IR\end{tabular}} & \textbf{\begin{tabular}[c]{@{}c@{}}BWV\\ PRS\end{tabular}} & \textbf{\begin{tabular}[c]{@{}c@{}}VS\\ IR\end{tabular}} \\ \hline
\multirow{2}{*}{\textbf{Mel vs Nevi}}     & \textbf{Traditional} & 2     & 1     & 1     & 1     & 1     & 2     & 2     \\
                                       & \textbf{Proposed}    & 1.81  & 1.36  & 1.20  & 0.87  & 0.93  & 2.35  &  1.66 \\ \hline
\multirow{2}{*}{\textbf{Mel vs Others}} & \textbf{Traditional} & 2     & 1     & 1     & 1     & 1     & 2     & 2     \\
                                       & \textbf{Proposed}    & 1.32  & 1.10  & 0.97  & 0.92  & 1.06  & 1.38  & 1.43  \\ \hline
\end{tabular}}
\end{table}

\subsection{Comparison to Traditional 7PCL Algorithm}

The approach presented in this paper is primarily inspired by the 7PCL clinical scoring mechanism for melanoma diagnosis, with key advancements introduced through dynamic weight assignment and automated diagnosis enabled by GD-DDW. These improvements not only enhance diagnostic performance and expand the applicability of the model but also generate data-driven weighting information that can provide valuable feedback to frontline clinicians. By leveraging real-world data for adaptive weighting, the proposed method addresses the inherent subjectivity in predefined scoring systems and improves the model’s interpretability.

To systematically evaluate these improvements, we compare the performance of the proposed method with the traditional 7PCL approach in two distinct diagnostic scenarios. The first follows the conventional application of 7PCL, distinguishing between Mel and Nevi, where the original scoring system has been widely validated. The second considers a more complex diagnostic setting in which Mel must be differentiated from a broader range of skin diseases. This latter scenario reflects the challenges encountered in real-world clinical applications, where dermatological conditions exhibit significant variability. By assessing performance across these scenarios, we aim to highlight the benefits of adaptive learning and dynamic weight assignment in improving diagnostic performance beyond traditional rule-based approaches. 

As shown in Table~\ref{tab:my-table-weights}, the traditional algorithm employs approximate categorization based on multivariate analysis \cite{argenziano1998epiluminescence}, assigning weights of 2 or 1 to major and minor attributes, respectively. In contrast, the method proposed in this study automatically generates adaptive weights based on the prediction scores of each category under different tasks. Specifically, in the classification task of differentiating Mel from Nevi, the weights of the major attributes range from 1.7 to 2.4, while the weights of the minor attributes range from 0.8 to 1.4, closely resembling traditional algorithms. However, when the task shifts to distinguishing Mel from four other skin diseases, the weights dynamically adjust to range from 1.3 to 1.5 for major attributes and 0.9 to 1.1 for minor attributes. 
 
\begin{figure}[t]
    \centering
    \subfloat[]{%
        \includegraphics[width=0.48\columnwidth]{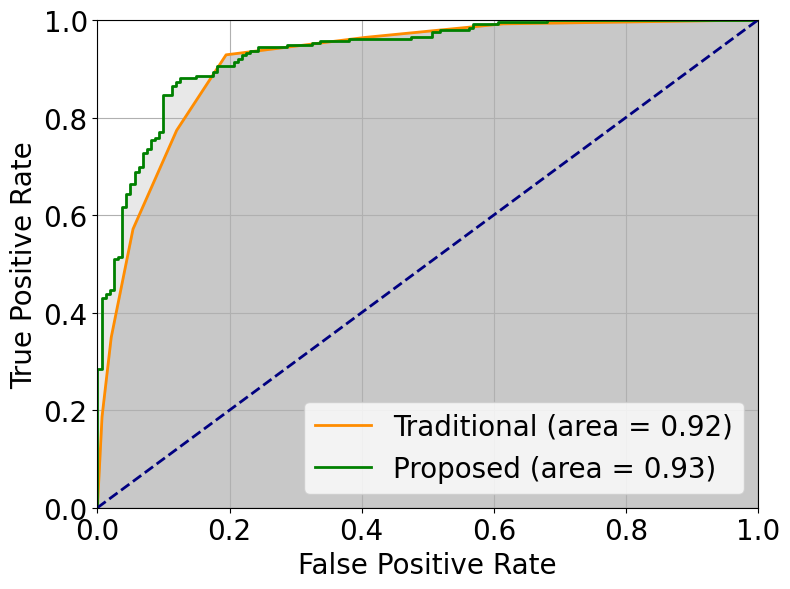}%
        \label{fig:curve_mm_mn}
    }  
    \subfloat[]{%
        \includegraphics[width=0.48\columnwidth]{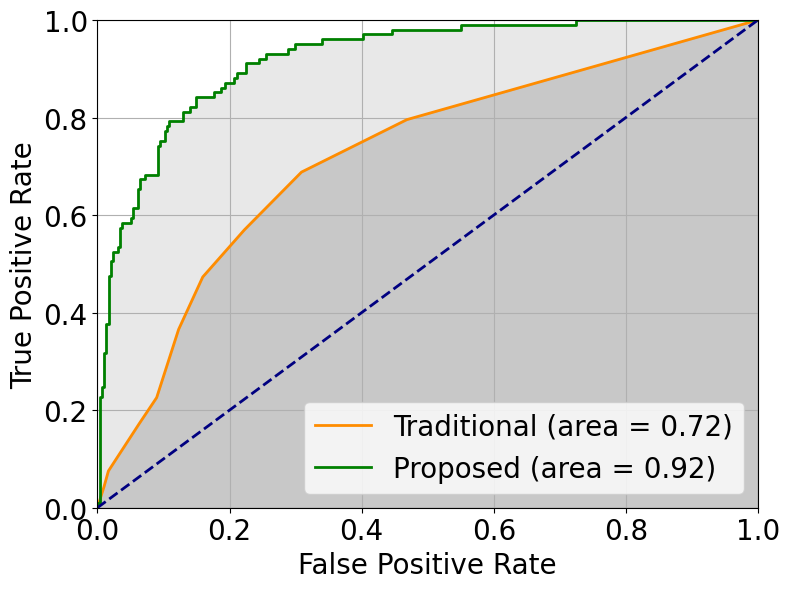}%
        \label{fig:curve_mm_others}
    }
    \caption{ROC comparison graph showcasing the performance differences between the proposed method and the traditional 7PCL algorithm across two classification tasks.(a): Classification of Mel and Nevi; (b): Classification of Mel and other diseases (MISC, Nevi, BCC, and SK)}
    \label{fig:roc_curve}
\end{figure}

 This dynamic adjustment extends the applicability of the 7PCL algorithm, originally designed to distinguish Mel from Nevi, beyond its initial scope. As shown in Fig.~\ref{fig:curve_mm_others}, the traditional algorithm experiences a significant drop in AUC, from 92\% to 72\%, when applied to more complex scenarios. In contrast, our proposed data-driven dynamic tuning approach maintains a high AUC value of 92\% while using the same features in the classification tasks of Mel vs. Nevi and Mel vs. others. It should be noted that the data-driven weight parameters in Table~\ref{tab:my-table-weights} are derived from model training and can be dynamically optimized as more data becomes available. These results indicate that although these attributes were originally designed for distinguishing Mel from Nevi, their diagnostic relevance extends beyond this binary classification. By enabling automated dynamic weight adjustment, the proposed method effectively preserves the value of these attributes in differentiating Mel from a broader range of skin diseases. This adaptability highlights the potential of leveraging clinically established features while optimizing their contribution based on task-specific requirements, thereby improving the model’s generalization in more complex diagnostic scenarios.

\subsection{Generalizability Analysis on the ISIC Datasets}

\begin{table}[t]
\centering
\footnotesize
\caption{Performance (AUC) of Proposed Method on ISIC Datasets.}
\label{tab:my-table-isic}
\resizebox{\columnwidth}{!}{%
\begin{tabular}{ll|cccc}
\hline
\textbf{Dataset} &
  \textbf{Attributes} &
  \textbf{\begin{tabular}[c]{@{}c@{}}Baseline\\ (ResNet-50)\end{tabular}} &
  \textbf{\begin{tabular}[c]{@{}c@{}}GRM-DC \\ \cite{fu2022graph} \end{tabular}} &
  \textbf{\begin{tabular}[c]{@{}c@{}}RemixFormer++ \\ \cite{xu2024remixformer++} \end{tabular}} &
  \textbf{\begin{tabular}[c]{@{}c@{}}CKTG\\ (Proposed)\end{tabular}} \\ \hline
\multirow{5}{*}{\textbf{ISIC2017}} & \textbf{Milia-like cysts} & 63.2          & 68.3          & 74.7 &  \textbf{75.4}  \\
                                   & \textbf{Negative network} & 77.9          & 84.1          & 86.8 &  \textbf{87.2}   \\
                                   & \textbf{Pigment network}  & 87.3          & 87.4          & 89.6 &  \textbf{91.2}   \\
                                   & \textbf{Streaks}          & \textbf{96.2}          & 93.9          & 95.6 &  95.2   \\
                                   & \textbf{Avg}              & 81.2          & 83.4          & 86.6 &  \textbf{87.3}   \\ \hline
\multirow{6}{*}{\textbf{ISIC2018}} & \textbf{Globules}         & 68.2          & 77.1          & 82.7 & \textbf{83.9} \\
                                   & \textbf{Milia-like cysts} & 60.1          & 68.8          & 74.3 & \textbf{76.4} \\
                                   & \textbf{Negative network} & 82.3          & 87.3          & \textbf{87.8 }& 87.6 \\
                                   & \textbf{Pigment network}  & 87.8          & 86.4          & 91.2 & \textbf{92.5} \\
                                   & \textbf{Streaks}          & 84.3          & 88.7          & \textbf{91.8} & 91.6 \\
                                   & \textbf{Avg}              & 76.5          & 81.7          & 85.6 & \textbf{86.4} \\ \hline
\end{tabular}}
\end{table}

Generalizability is a key factor in assessing a method's clinical applicability. To evaluate this, we tested our approach on the ISIC 2017 and ISIC 2018 datasets and compared it with state-of-the-art methods. While these datasets contain only dermoscopic images, they encompass a diverse range of skin conditions, including melanoma, Nevi, and SK, and provide attribute annotations beyond the 7PCL framework, such as milia-like cysts. This evaluation assesses the model's ability to generalize beyond 7PCL-specific attributes while accounting for dataset variations, demonstrating its robustness and adaptability to real-world dermatological diagnosis.

As shown in Table~\ref{tab:my-table-isic}, we compared the predictive performance of our method against the baseline model (ResNet-50) and SOTA approaches (GRM-DC and RemixFormer++). Our method achieved the highest average AUC across both datasets, with 87.3\% on ISIC 2017 and 86.4\% on ISIC 2018. It also attained the highest classification performance in six out of nine attributes, further demonstrating its robustness. These results highlight the effectiveness of our method in extracting representative features and modeling attribute interactions, enhancing predictive performance for broader dermatological applications.

\section{Discussion}
In this study, we proposed a novel approach to melanoma diagnosis, integrating clinical empirical knowledge, diagnostic procedures, and GCNs. Specifically, we improved the causal learning of melanoma-associated attributes to construct an optimized topological graph structure. By incorporating computational logic from clinical algorithms, our method enhances interpretability and clinical acceptance by strengthening the correlation between diagnostic outcomes and imaging datasets. Additionally, several key aspects identified during this study warrant further discussion.

\subsection{Multi-Order Node Associations and Their Impact}
Our results demonstrate that multi-order node associations, introduced through ARFP, play a crucial role in melanoma detection performance. Beyond the main results presented in the results section, additional experimental evaluations reveal a consistent improvement in AUC as the order \( k \) increases: 87.1\% at \( k=1 \), 90.6\% at \( k=2 \), and 92.0\% at \( k=3 \), indicating that expanding the receptive field effectively captures more informative attribute relationships. However, at \( k=4 \), AUC slightly declines to 91.8\%, suggesting a performance plateau. Given that the 7PCL graph comprises only 7 nodes, further increasing \( k \) reduces the number of meaningful paths, leading to information dilution rather than enrichment.

These observations highlight the importance of selecting an optimal order \( k \) to balance local feature aggregation and global information propagation. Future work may explore dynamic receptive field adjustment or adaptive graph construction based on dataset-specific connectivity patterns. Additionally, incorporating more informative labels could help mitigate the limitations imposed by the small number of nodes, potentially enhancing graph-based melanoma classification performance.

\subsection{Semantic Information and Its Role in Graph Learning}

Existing research on graph-based melanoma detection has primarily focused on link relationships between nodes, while the semantic content of node labels has often been overlooked \cite{tang2023graph,fu2022graph,tong2020digraph}. To the best of our knowledge, this study is the first to integrate dynamic semantic information into both feature extraction and the graph learning process, enabling full utilization of diagnostic knowledge.

As shown in Table~\ref{tab:my-table-encoding}, we evaluated multiple semantic embedding methods, including one-hot encoding \cite{rumelhart1986learning}, GloVe \cite{pennington2014glove}, Word2Vec \cite{church2017word2vec}, and BERT \cite{devlin2018bert}. Each method differs in its ability to capture contextual relationships among node labels, with dynamic embeddings offering greater adaptability. 

Results indicate that models incorporating semantic embeddings outperform those without them, with BERT achieving the highest AUC values of 92.0\% for melanoma classification and an average AUC of 88.6\% across all attributes. However, the inclusion of dynamic embeddings comes at the cost of increased computational time, particularly with BERT requiring 84 minutes of training time compared to 56-63 minutes for static embeddings. Despite this trade-off, the improved diagnostic performance demonstrates the effectiveness of integrating contextual knowledge into graph-based melanoma detection.

These findings suggest that leveraging dynamic embeddings allows the model to capture higher-order semantic relationships, ultimately enhancing classification performance. Future research could explore hybrid approaches, balancing computational efficiency with embedding effectiveness for large-scale dermatological datasets.

\begin{table}[t]
\footnotesize
\caption{Comparison of Different Semantic Embedding Methods.}
\label{tab:my-table-encoding}
\resizebox{\columnwidth}{!}{%
\begin{tabular}{cccccccc}
\hline
\multirow{2}{*}{} & \multirow{2}{*}{\textbf{\begin{tabular}[c]{@{}c@{}}Semantic \\ Information\end{tabular}}} & \multirow{2}{*}{\textbf{\begin{tabular}[c]{@{}c@{}}Dynamic \\ Embedding\end{tabular}}} & \multicolumn{2}{c}{\textbf{Training time}} & \multicolumn{2}{c}{\textbf{AUC}} \\
 &  &  & \textbf{Encoding} & \textbf{Total} & \textbf{Mel} & \textbf{Avg} \\ \hline
\textbf{One-Hot} & $\times$ & $\times$ & 0.3s & 63 mins & 86.8 & 81.1 \\
\textbf{GloVe} & $\checkmark$ & $\times$ & 1s & 61 mins & 89.1 & 83.7 \\
\textbf{Word2Vec} & $\checkmark$ &$\times$  & 1s & 56 mins & 88.7 & 84.0 \\
\textbf{BERT} & $\checkmark$ & $\checkmark$ & 3s & 84 mins & 92.0 & 88.6 \\\hline
\end{tabular}%
}
\end{table}

\subsection{Clinical Implications and Adaptive Weighting for 7PCL}
From a clinical perspective, the 7PCL algorithm was originally developed to assist dermatologists in distinguishing Mel from Nevi based on dermoscopic features. However, adapting it into a modern computer-aided diagnosis (CAD) system that encompasses a broader range of similar-looking skin lesions and integrates both clinical and dermoscopic features presents significant challenges. One key limitation is that the original scoring system, which assigns a weight of 2 to major features and 1 to minor features, was designed specifically for Mel and Nevi differentiation and may not generalize well to more diverse diagnostic scenarios \cite{argenziano1998epiluminescence,walter2013using,islam2024p028}.

To address these limitations, this study introduces a data-driven approach that dynamically adjusts attribute weights based on the classification task. By refining the weighting mechanism, the 7PCL algorithm can adapt to different dermatological conditions while maintaining its clinical interpretability. This adjustment allows the system to better align with real-world diagnostic needs, making it more applicable in various clinical settings and improving its utility as a decision-support tool for medical professionals.

\subsection{Potential for Integrating Foundation Models}
Foundation models, particularly large-scale multimodal architectures, have demonstrated remarkable adaptability across various domains, including dermatology. These models, pretrained on vast and diverse datasets, offer strong feature representations that can be fine-tuned for specific medical tasks, including melanoma diagnosis. Given their ability to capture complex hierarchical patterns and cross-modal interactions, integrating foundation models into our framework could further enhance its robustness and generalization. Specifically, vision-language models like SkinGPT-4 and pathology foundation models have shown promise in leveraging both textual clinical knowledge and visual cues, potentially enriching the diagnostic reasoning process in automated systems \cite{yan2024general,zhou2024pre,wang2024pathology,bluethgen2024vision}.

Incorporating foundation models into our approach could provide several advantages. First, they could refine the dynamic weighting mechanism by learning from broader datasets, improving the adaptability of the 7PCL-based scoring system beyond its original scope. Second, their self-supervised learning capabilities could mitigate the limitations of annotated medical datasets, allowing for improved feature extraction even in data-constrained environments. Lastly, by integrating pretrained multimodal embeddings, our framework could benefit from enhanced attribute interactions and context-aware diagnostics, making it more interpretable and clinically applicable. Future work will explore the feasibility of integrating foundation models to further advance automated melanoma diagnosis.

\section{Conclusion}

In this study, we present an artificial intelligence-enhanced 7-Point Checklist (7PCL) method for melanoma detection, designed to integrate a clinically-informed topology that mirrors the diagnostic approach of dermatologists. This approach addresses the limitations of the traditional 7PCL, which has been primarily restricted to differentiating Mel from Nevi, by enabling automated adaptation to diverse diagnostic scenarios. Our method efficiently extracts visual features using an architecture that combines single-modal and cross-modal attention mechanisms, capturing subtle relationships within and across modalities. By incorporating directed multi-order relationships and gradient prediction structures, our approach dynamically assigns weighted metrics to each attribute. This allows the algorithm to adapt to varying diagnostic tasks, such as distinguishing Mel from a broader range of skin conditions with overlapping visual characteristics. This enhanced weighting system not only improves detection performance but also provides dermatologists with a more intuitive and interpretable framework for attribute analysis. By aligning the algorithm’s design with clinical workflows, the proposed method could enhance both the effectiveness and acceptance of AI-assisted diagnosis in real-world settings.

\bibliographystyle{IEEEtran}  
\bibliography{ref}
\vfill
\end{document}